\PassOptionsToPackage{table}{xcolor}
\documentclass{article}

\usepackage{multirow}
\usepackage{colortbl}
\usepackage{graphicx}
\usepackage{makecell}
\usepackage{fullpage}
\usepackage{amsmath}
\usepackage{amssymb}
\usepackage{booktabs}
\usepackage{xcolor} 
\usepackage{enumitem}
\usepackage{multirow}
\usepackage{algorithm}
\usepackage{algpseudocode}
\usepackage[numbers]{natbib}
\usepackage{wrapfig}

\def\x{{\bf x}}

\def\0{{\bf 0}}
\def\1{{\bf 1}}

\definecolor{darkgreen}{rgb}{0.3, 0.75, 0.3}
\definecolor{lightred}{rgb}{1.0, 0.5, 0.5}
\definecolor{lightgrey}{rgb}{0.9, 0.9, 0.9}
\newcommand{\cmark}{\textcolor{darkgreen}{\checkmark}}%
\newcommand{\xmark}{\textcolor{lightred}{$\times$}}%

\definecolor{lightblue}{rgb}{0.21,0.49,0.74}
\usepackage[pagebackref,breaklinks,colorlinks,allcolors=lightblue,hypertexnames=false]{hyperref}
\usepackage{stfloats}
\usepackage[capitalize]{cleveref}

\title{Observing Quietly, Assisting Precisely: \\Detecting Personal Items Based on Video Observation}

\author{Yechi Ma \quad Nan Deng
\quad
Wei Hua
\quad
Shu Kong \\
{\em website: \url{https://elowendeng.github.io/PID/}}
}
\date{}

\begin{document}
\maketitle

\begin{abstract}\noindent
An emerging capability for visual systems is observing users' daily lives to help locate misplaced personal items.
We formalize this task as Personal Item Detection (PID).
Given unlabeled daily-life reference videos,
PID algorithms detect individual users and their personal items to help users detect target items in cluttered environments based on fuzzy, user-friendly, natural-language (FUN) queries, e.g., ``where is my wallet?'' asked by Sophie, who is one of multiple users.
To support this research, we introduce Video Observation for PID (VOPID),
a dataset featuring daily interaction videos alongside challenging test images of personal items in complex scenes.
PID presents three key challenges: 
(1) identifying multiple users and inferring object ownership from unconstrained videos,
(2) memory bank design to store mined knowledge,
and (3) performing cross-modal matching between FUN queries and the stored knowledge. 
To address these challenges,
we propose a modular pipeline centered on a memory bank, allowing evaluation of both classic computer vision methods and modern foundation models for object detection, tracking, and matching.
Extensive experiments reveal critical limitations of
modern AI agents, which achieve only 4.0 mAP;
however, strategically integrating foundation models within the modular pipeline boosts the performance to 30.3 mAP.
The results establish strong baselines and outline future research directions.
\end{abstract}

\begin{figure*}[h]
\centering
\includegraphics[width=\linewidth]{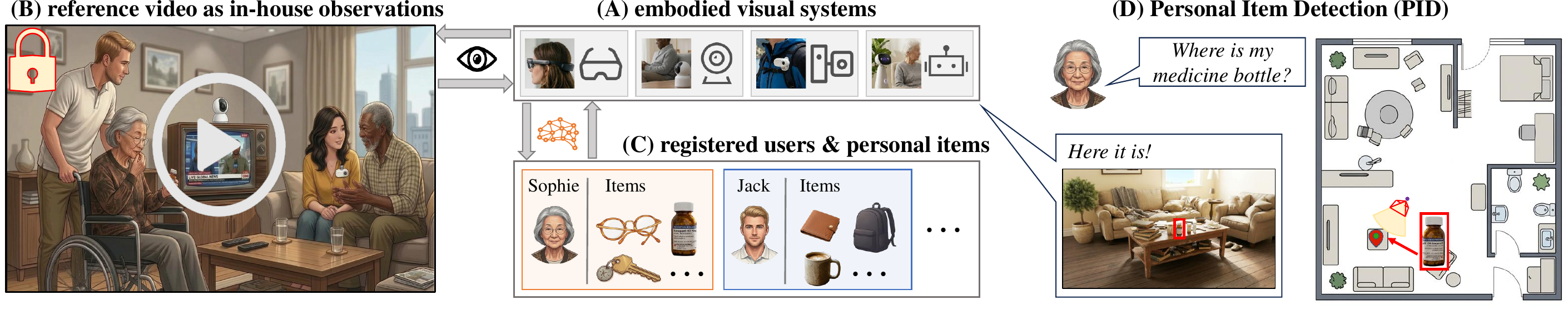}
\vspace{-9mm}
\caption{\small
\textbf{Overview of Personal Item Detection (PID)}.
\textbf{(A)}
Embodied visual systems (e.g., smart glasses, indoor cameras,
wearable AI, and assistive robots)
continuously observe daily user activities and interactions.
\textbf{(B)}
These observations naturally manifest as privacy-sensitive, unlabeled,
unconstrained video streams recorded within private residences.
\textbf{(C)}
To effectively assist individuals,
these systems should parse such videos offline to identify users,
detect all possible personal items, establish ownerships,
and store this knowledge in a memory bank.
\textbf{(D)}
During inference,
the systems leverage this memory bank to help users locate personal items
in cluttered environments based on
\textit{fuzzy, user-friendly, natural-language (FUN)} queries,
such as ``\textit{where is my medicine bottle?}''
asked by Sophie (a user among multiple ones).
Solutions for PID hold profound practical significance,
enhancing daily living and providing assistance to users in need \cite{kwon2024accesslens, arabi2026eurexa}.
}
\label{fig:overview}
\end{figure*}

\vspace{-4mm}
\section{Introduction}
Visual systems operating in private residences continuously observe users,  activities, interactions, and surrounding environments via video streams \cite{arabi2026eurexa} (Fig.~\ref{fig:overview}A\&B)
to deliver a personalized service termed \textit{Personal Item Detection (PID)}:
helping individuals locate belongings based on fuzzy, user-friendly, natural-language (FUN) queries \cite{hatori2018interactively}, such as ``where is my wallet?'' asked by Sophie, who is one of multiple users.

While PID is a novel formulation,
prior studies have explored related aspects of personalized visual perception, including instance detection and tracking \cite{Shen2023AHD, barbato2024cross, Shen2025SolvingID, camuffo2026mocha, lee2025bring, zhao2024instance},
instance manipulation and navigation \cite{barsellotti2024personalized, lee2025bring, wu2026revisitingmonocularinhandobject},
and personalized QA \cite{alaluf2024myvlm, nguyen2024yo, pham2025plvm, hao2025rap}.
However, these works rely on overly simplified settings,
such as manual pre-enrollment of users and their items of interest,
omitting FUN queries (Fig.~\ref{fig:overview}D) by multiple users.
Other works investigate conversational enrollment strategies, prompting users to answer questions like ``whose cup is this?'' \cite{hashimoto2026toward} or describe items in detail (e.g., ``the shoes I got for that hiking trip last summer, which I always wear for long walks now'') \cite{lee2026personalizing}.
Nevertheless, they still impose significant interaction burdens on users.
In contrast, practical PID achieves frictionless enrollment: 
agents quietly observe daily life offline to safeguard privacy, recognize and disambiguate items,
autonomously associate people with their belongings,
and store this knowledge locally (Fig.~\ref{fig:overview}C) to resolve users' FUN queries.

To support PID research,
we introduce \emph{Video Observation for PID (VOPID)},
a benchmark using a monitoring camera\footnote{PID is a general task that can be executed across diverse embodied agents, including mobile robots, egocentric wearable cameras, and stationary surveillance cameras. Our work focuses on the latter setting.} to capture rich human-object and human-human interactions among multiple users in living and working environments. 
VOPID comprises a validation set and a test set,
totaling 41 reference videos and 519 FUN queries by 9 individuals alongside test images.
All participating users provided informed consent for the release of their personally identifiable information to foster PID research.

To address PID,
we identify three key challenges.
(1) \emph{Knowledge mining from unlabeled, unconstrained videos:} 
As reference videos lack annotations and do not provide explicit supervision regarding individuals and belongings,
PID algorithms must comprehensively parse the videos in terms of person detection, object detection, and ownership recognition.
(2) \emph{Memory design:} 
Extracted knowledge must be structured and stored effectively to serve as a reliable reference for PID algorithms to process FUN queries.
(3) \emph{Query-instance matching:} 
Given a test image and a FUN query,
the algorithm must detect candidate objects that match the query while explicitly abstaining from returning any detections when the target item is invisible or absent.

While we evaluate end-to-end AI agents in directly detecting personal items based on FUN queries and video reference,
we also introduce a conceptually simple modular pipeline, with each module addressing one of the three challenges.
Within each module,
we rigorously compare both conventional methods and modern foundation models.
For knowledge mining in videos,
we study classic computer vision methods for object detection and tracking, and foundation models.
We further explore different memory bank designs to store mined knowledge, including explicit JSON files \cite{park2023generative} and implicit KV cache from AI agents \cite{yang2026keep}.
Based on these memory structures,
we develop methods to detect target items from test images based on FUN queries.
The variety of explored methods yields rich insights:
AI agents struggle with PID, only achieving 4.0 mAP,
whereas strategically integrating foundation models within our modular pipeline achieves 30.3 mAP.
These results establish strong baselines and outline future research directions.

We make three primary contributions:
\begin{enumerate}[leftmargin=15pt, topsep=0pt, itemsep=5pt,parsep=-2pt]
    \item \textit{Problem formulation.}
    We formulate the problem PID as a concrete personalized visual perception task that drives the future development of embodied agents.
    
    \item \textit{Benchmark construction.} 
    We introduce VOPID, 
    a new benchmark that fosters research in PID and serves as an effective platform to evaluate contemporary AI agents in personalized perception and reasoning. 
    
    \item \textit{Methodological development.} 
    We propose a conceptually simple, modular pipeline to solve PID,
    enabling comprehensive evaluation of both traditional methods and foundation models across modules and yielding insightful results that inform future research.
\end{enumerate}

\section{Related Work}
\label{sec:related-works}

\textbf{Object detection at different granularities.}
Category-level object detection has been extensively studied \cite{Felzenszwalb2010ObjectDW, Ren2015FasterRT, Carion2020EndtoEndOD}.
The detectors detect all objects belonging to predefined categories of objects.
Recent work extends category-level detection to open-vocabulary detection \cite{li2022grounded, Minderer2023ScalingOO, liu2024grounding, wang2026locateanything},
which develops foundational detectors to detect objects and their parts based on language queries.
However,
category-level and open-vocabulary detection do not necessarily work in personalized scenarios,
which require detecting specific object instances.
Consequently, instance-level detection is proposed to detect specific object instances based on reference images  \cite{Dwibedi2017CutPA, Mercier2021DeepTO, Shen2023AHD, barbato2024cross, Paramonov2024SwissDE, camuffo2026mocha, Shen2025SolvingID}.
However, reference images do not come for free but require users to pre-enroll the target instances \cite{lee2025bring}, e.g., taking photos of the instances of interest beforehand.
To reduce the pre-enrollment burden,
some methods develop conversational enrollment strategies by prompting users to answer questions like ``whose cup is this?''  \cite{hashimoto2026toward} or describe items in detail (e.g., ``the shoes I got for that hiking trip last summer, which I always wear for long walks now'') \cite{lee2026personalizing}.
Nevertheless, 
they still impose interaction burdens on users.
In contrast,
our formulated problem Personal Item Detection (PID) requires algorithms to detect all users from videos toward auto-enrollment.

\textbf{Personalized visual perception} has been explored via different tasks, including 
instance detection \cite{Shen2023AHD, barbato2024cross, Shen2025SolvingID},
instance tracking \cite{xu2023my, zhao2024instance},
image retrieval \cite{Das2025TrainingFreePV, wang2026generating, yang2026beyond},
instance navigation \cite{lee2026personalizing, barsellotti2024personalized},
instance manipulation \cite{lee2025bring, wu2026revisitingmonocularinhandobject}, 
and
personalized QA \cite{alaluf2024myvlm, nguyen2024yo, pham2025plvm, seifi2026ego}.
Most existing studies treat instance-level perception as a proxy for personalized perception, where individuals specify object instances for models to detect, track, and search.
However, this proxy overlooks the non-trivial burden of enrolling target instances,
as users must manually supply reference photos.
Furthermore, 
prior research has largely neglected the challenges of multi-user personal instance enrollment.
\cite{yeh2023meta} records users' historical descriptions to associate them with target objects, thereby establishing object ownership;
\cite{hashimoto2026toward} actively prompts users to confirm ownerships of detected objects;
\cite{huber2026cool} explores ownership
learning from unlabeled videos.
We complement this line of work by providing
unlabeled, unconstrained videos that capture multiple users
and intensive human-human and human-object interactions.
It requires PID algorithms to establish object ownership
from these videos and localize personal items from user queries.
On the other hand,
to achieve personalization, existing methods develop external memory banks to store personal information.
For example,
some methods store images of target objects as visual references
\cite{barbato2024cross, zhao2024instance, Shen2025SolvingID},
whereas others create human-readable files storing both text and images for users and target objects \cite{nie2026personavlm, Das2025TrainingFreePV, hao2025rap, hong2026tame, lee2026personalizing}.
Building upon the observation that modern Multimodal Large Language Models (MLLMs) utilize a KV cache to store intermediate states, such as dialogue history and user prompts \cite{zhang2023h2o, shi2024keep, brandon2024reducing, xiao2024efficient, li2024snapkv, liu2024cachegen, liu2024deepseek},
we show that this KV cache can be repurposed as an external memory bank for PID.
We systematically evaluate this memory design and traditional human-readable files for PID.

\begin{wraptable}{r}{0.55\linewidth}
\centering
\small
\vspace{-5mm}
\caption{\small
\textbf{Comparison of VOPID and related datasets.} 
While emphasizing personalization in visual perception, 
existing datasets do not necessarily 
(1) admit FUN queries by multiple users,
(2) provide reference videos,
(3) require ownership learning for auto-enrolling users and objects,
(4) contain socially intensive interactions,
or
(5) require actionable bounding box output.
The comparison highlights the uniqueness and challenge of both the VOPID benchmark and the PID problem.
}
\label{tab:benchmark_comparison}
\resizebox{0.96\linewidth}{!}{%
\setlength{\tabcolsep}{0.3em}
\begin{tabular}{@{}llcccccc@{}}
\toprule
\makecell{Dataset} & \makecell{primary\\task} & \makecell{FUN\\query} & \makecell{video\\ref.} & \makecell{owner.\\reason.} & \makecell{social\\intense} & \makecell{act.\\box} \\

\midrule
MyVLM \cite{alaluf2024myvlm} &  VQA & \xmark & \xmark & \xmark & \xmark  & \xmark \\
Yo'LLaVA \cite{nguyen2024yo} &  VQA & \xmark & \xmark & \xmark & \xmark & \xmark \\
PLVM \cite{pham2025plvm} &  VQA & \xmark & \xmark & \xmark & \xmark & \xmark\\
RAP \cite{hao2025rap} &  VQA & \xmark & \xmark & \xmark & \xmark  & \xmark\\
MyEgo \cite{xiao2026ego} & VQA & \xmark & \cmark & \cmark & \cmark & \xmark \\
HR-InsDet \cite{Shen2023AHD} & Ins. Det. & \xmark & \xmark & \xmark & \xmark & \cmark\\
POD~\cite{barbato2024cross} & Ins. Det.  & \xmark & \xmark & \xmark & \xmark & \cmark  \\
IT3DEgo \cite{zhao2024instance} & Ins. Track. & \xmark & \cmark & \xmark & \xmark & \cmark \\
VLABench \cite{lee2025bring} & Ins. Manip. & \xmark & \xmark & \xmark & \xmark & \xmark \\
COOL \cite{huber2026cool} & Ins. Nav. & \xmark & \cmark & \cmark & \xmark & \xmark  \\
Ego4D-VQ2D~\cite{grauman2022ego4d} & Temp. Loc. & \xmark & \cmark & \xmark & \xmark & \cmark \\
EgoLife~\cite{yang2025egolife} &  VideoQA & \xmark & \cmark & \xmark & \cmark & \xmark \\
EgoMemory~\cite{tang2026egomemory} &  Video Ret. & \xmark & \cmark & \cmark & \xmark & \xmark \\

\midrule
\textbf{VOPID (Ours)} & \textbf{PID} & \cmark & \cmark & \cmark & \cmark & \cmark \\
\bottomrule

\end{tabular}
}
\vspace{-3mm}
\end{wraptable}

\textbf{Knowledge mining from videos.}
Natural videos can capture rich interactions among people.
Early work mines human-object interactions from daily-life vlogs \cite{fouhey2018lifestyle} and hand-object contact and motion from web videos \cite{shan2020understanding, shan2021cohesiv}.
These studies establish interaction cues as a means of extracting object-centric knowledge from unconstrained videos. 
Extending interactions,
\cite{hashimoto2026toward} explores ownership understanding with a human-in-the-loop strategy: models actively prompt users to confirm ownerships of detected objects.
\cite{huber2026cool} aims to infer object ownership from videos without human intervention; yet, each of their videos captures a simple scene, sparse objects, and a single user.
We release a new dataset that captures rich human-human and human-object interactions in complex scenes,
requiring algorithms to infer ownership from such videos.
On the other hand, video MLLMs, such as Video-ChatGPT~\cite{maaz2024videochatgpt}, have emerged that enable open-ended descriptions and question answering over video input.
While underexplored, video MLLMs can be leveraged to reason about object ownership.
Other than ownership mining from videos,
the problem PID also emphasizes FUN queries at test time, when a specific user asks ``where is my wallet?''
That said, PID algorithms must extract not only ownership relations but also users' identities and objects' appearances and descriptions.

\section{Problem Formulation and Benchmark}
\label{sec:PID}
We begin with the formulation of PID.
Then, we introduce the VOPID benchmark including both curation details and ethical aspects.
Lastly, we establish evaluation protocols.

\subsection{Problem Definition}
Given a tuple input $(u, q, I)$,
comprising a user's portrait\footnote{Identifying the user can practically leverage both visual re-ID and acoustic cues.
To focus on core methodology development, we simplify the re-ID process by providing user profile images alongside the FUN queries.} $u$, 
their FUN query $q$ (e.g., ``Where is my cup?''),
and a test image $I$,
PID aims to detect the target instances in $I$ based on $u$ and $q$.
While PID does not provide annotations to support training (in light of privacy-sensitive application scenarios),
the algorithms have access to unlabeled, unconstrained videos as references,
from which they can mine knowledge, e.g., users' identities, all possible objects and their ownerships.
The algorithms can leverage the mined knowledge for PID.

\subsection{Dataset Curation}
\textbf{Desiderata.}
To support PID research,
we curate our VOPID benchmark dataset with the following desiderata.
\Cref{tab:benchmark_comparison} is a high-level comparison with existing datasets, demonstrating the uniqueness and value of VOPID.

\begin{itemize}[leftmargin=12pt, topsep=1pt, itemsep=3pt, parsep=0pt]
    \item 
    \textit{FUN queries.}
    Users naturally communicate with fuzzy terms  (e.g., ``where is my coffee mug?'') instead of detailed visual descriptions.
    We collect real FUN queries from multiple users in shared living or working spaces.

    \item
    \textit{Intensive interactions.} 
    Real environments involve multiple users socially interacting with each other and diverse objects.
    Our videos capture this complexity and highlight the challenge of ownership reasoning from unlabeled, unconstrained videos.

    \item
    \textit{Distractors.} 
    Everyday objects, e.g., coffee mugs, of different owners exhibit subtle visual variations.
    We collect videos containing diverse distractors, including same-category items associated with different users, to evaluate a method's ability to reason over item ownership.

    \item \textit{Actionable localization.}
    Many embodied agents require precise localization of target objects (for grasping).
    We collect test images with manual box annotations to benchmark PID methods w.r.t 2D detection performance.

    \item \textit{Video length.}
    While long videos are available in real-world settings, their substantial computational costs make iterative development of PID methods less efficient.
    Hence, we collect short videos as references.

    \item
    \textit{Indoor security cameras}
    provide a broader field of view capturing user-object interactions,
    in contrast to the narrower, egocentric perspectives of wearable or mobile robotic cameras.
    This setup opens promising avenues for future research, e.g.,  multi-camera agents.
\end{itemize}

\begin{wrapfigure}{r}{0.55\linewidth}
  \centering
  \vspace{-5mm}
  \includegraphics[width=\linewidth]{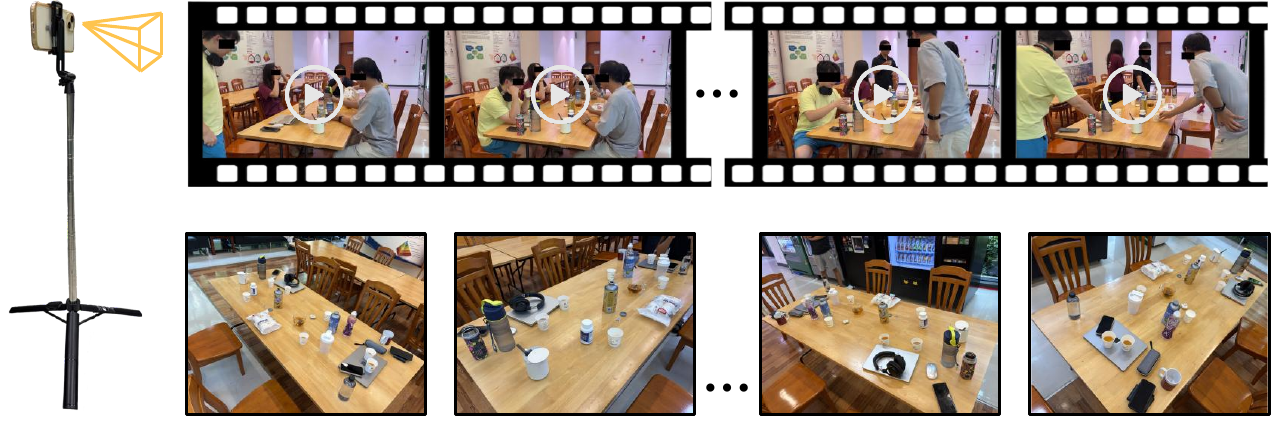}
  \vspace{-9mm}
  \caption{\small \textbf{Data collection} via a tripod-mounted smartphone.}
  \label{fig:data_collection}
  \vspace{-3mm}
\end{wrapfigure}

\textbf{Video collection.}
We used a smartphone mounted on a high tripod to record videos (Fig.~\ref{fig:data_collection}).
Nine participants contributed recordings of their daily interactions across shared offices, living rooms, meeting rooms, and recreation lounges. 
From raw videos, we select shorter clips (ranging from 20s to 4min) as the final references.

\textbf{Test data.}
We use the same device to capture test images of the scene under varying  viewpoints and distances (Fig.~\ref{fig:data_collection}).
While these variations are less frequent with fixed security cameras,
introducing them yields a more challenging benchmark that prevents PID methods from memorizing previous object locations.
Participants annotate their belongings in each test image using bounding boxes and  provide corresponding FUN queries.
We then combine their profile images, FUN queries, and test images to form the final test set.
Notably, we also incorporate negative queries referencing instances absent from the test images,
evaluating a PID algorithm's ability to say ``I don't know.''

\textbf{Ethics.}
Our dataset contains personally identifiable information (PII),
such as facial images and audio recordings.
All participants granted explicit consent for the release of their PII to support PID research.
This study was reviewed and approved by our IRB.
The dataset will be released under a CC BY-NC-ND license. 

\subsection{Evaluation Protocol}
\textbf{Tracks.}
We present two tracks w.r.t user registration:
\begin{itemize}[leftmargin=12pt, topsep=1pt, itemsep=3pt, parsep=0pt]
    \item \textit{User pre-registration (UPR)} 
    simulates the scenario in which users register themselves when deploying a visual system by uploading their personal information such as profile images.
    Then, PID algorithms need to re-identify users from videos according to registered users and only serve them at test time.
    
    \item \textit{User auto-registration (UAR)} requires PID algorithms to automatically discover and register different users and serve them later at test time.
\end{itemize}

\textbf{Data Splits.}
VOPID is split into validation (10 videos, 130 queries) and test (31 videos, 389 queries) sets, with no shared videos (refer to Suppl.~Table~\ref{tab:supp-dataset-statistics} for partition details). 
Hyperparameters and prompts are tuned on validation and fixed for test evaluation.

\textbf{Diagnostic subsets of test data.}
Test queries are partitioned across three operational dimensions:
\begin{enumerate}[leftmargin=12pt, topsep=1pt, itemsep=2pt, parsep=0pt]
    \item \textit{Instance size.}
    We categorize test queries w.r.t the ratio $r$ of the target-instance area to the image area:
    small ($r$$<$1$\%$; 218 queries),
    medium ($1\%$$\le$$r$$<$5$\%$; 112 queries),
    and large ($r$$\ge$5$\%$; 25 queries).

    \item \textit{Occlusion severity.}
    We categorize test queries w.r.t  
    the manually annotated fraction of each target that is occluded, denoted by $o$:
    unoccluded ($o$$=$0; 239 queries),
    partially occluded (0$<$$o$$<$0.5; 87 queries), 
    and heavily occluded ($o$$\ge$0.5; 29 queries).
    Absent queries are excluded from this group.

    \item \textit{Target absence.}
    We specifically evaluate on  queries asking about instances absent from test images.
    There are 34 such queries.
\end{enumerate}

\textbf{Metrics.}
We evaluate performance using mean Average Precision ($\text{mAP}$), averaged across all test queries and IoU thresholds from 0.50 to 0.95 with a step size of 0.05.
Conventionally,
mAP is calculated as the area under the precision-recall curve by sweeping confidence scores.
However, real-world deployment typically requires a single operating point rather than confidence sweeping.
Therefore, 
we decouple post-processing (e.g., confidence thresholding) from evaluation, leaving the former to the algorithms themselves and focusing strictly on their final detections.
That said, our mAP measures correctness.
This formulation naturally accommodates instance-absent queries and the outputs from MLLMs, which generally do not provide confidence scores.
See Suppl.~\S\ref{supplsec:evaluation} for metric and matching rules.

\section{Methodology Development}
\label{sec:method}
\textbf{A modern method based on AI.} 
We leverage modern AI agents and design a baseline, which
prompts an MLLM capable of video analysis (e.g., Qwen 3.8 \cite{qwen3.8}) to directly 
output the bounding boxes of target instances, or ``not found'' if none are found, 
conditioned on the user's profile image, the FUN query, the reference video, and the test image.
Notably, 
because this method ingests the entire reference video for each FUN query,
it treats the videos as an external memory bank.
Consequently, it requires no user registration and can be applied across UPR and UAR settings.
However, despite consuming a massive number of tokens,
it yields poor PID performance (\Cref{tab:pidove-main}).
To diagnose this failure and improve performance,
we introduce a modular pipeline next.

\textbf{A classic method.}
From first principles,
we use classic computer vision techniques to address the three key challenges of PID:
(1) mining user-item knowledge from reference videos,
(2) structuring the memory bank to store this knowledge,
and (3) querying the memory bank to localize target instances from a user's FUN query.
This formulation naturally leads to a modular pipeline.
Below, we instantiate each module using dedicated computer vision approaches and subsequently demonstrate how they can be enhanced with MLLMs and  foundation models.

\begin{figure}[t]
  \centering
  \includegraphics[width=0.94\linewidth]{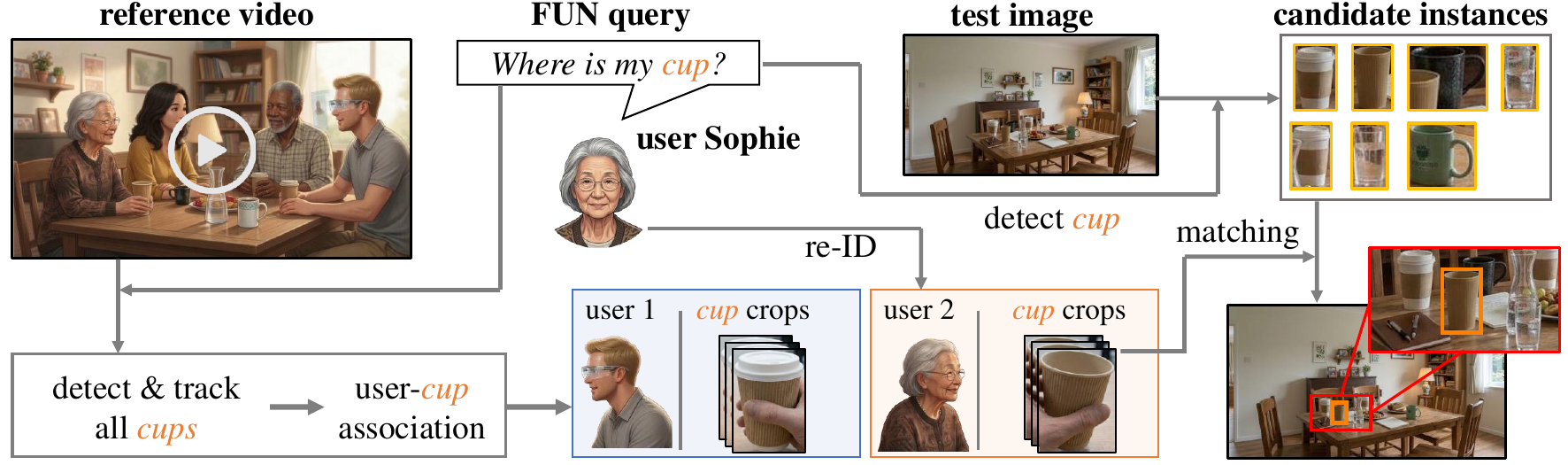}
\vspace{-3mm}
  \caption{\small 
  \textbf{A modular pipeline built with classic methods.}
  It broadly adopts a proposal-matching paradigm:
  detecting candidate instances in both the reference video and the test image,
  associating candidate instances with interacting users,
  matching those users against the profile image of Sophie (the requester),
  and matching Sophie's retrieved cup against candidate cups in the test image.
  Refer to \S\ref{sec:pid-traditional} for details.
  }
  \label{fig:pid_cv_baseline}
\vspace{-0mm}
\end{figure}

\subsection{A Modular Pipeline with Classic Methods}
\label{sec:pid-traditional}
Fig.~\ref{fig:pid_cv_baseline} depicts this pipeline, which integrates proposal detection, person re-ID, and instance-level matching.
Specifically,
given a FUN query such as ``Where is my \textit{cup}?'' asked by Sophie,
the pipeline parses the query and targets the object category \textit{cup}.
It detects all individuals and \textit{cups} \cite{liu2024grounding,wang2026locateanything} in the reference video,
tracks the \textit{cups} across frames \cite{zhang2022bytetrack},
and associates each \textit{cup} with the spatially closest user.
Meanwhile, it detects all candidate \textit{cups} in the test image.
By matching the detected users against Sophie's profile image via person re-ID \cite{deng2019arcface}, 
it retrieves crops of Sophie's \textit{cup} in video frames.
It uses these crops to match against the test-image detections via instance-level matching \cite{Shen2025SolvingID, zhao2024instance} using off-the-shelf features \cite{oquab2023dinov2}.
The bounding boxes of all target \textit{cups} with matching scores
above a predefined threshold serve as the final PID output.

This modular classic pipeline applies to both UPR and UAR settings 
and treats raw reference videos directly as an unindexed memory bank.
However, 
it couples video-based knowledge mining with each FUN query,
hence it requires full video processing for each query,
making the method computationally impractical.
To overcome this limitation and boost PID performance,
we next enhance each module in the pipeline using MLLMs and foundation models.

\subsection{Knowledge Mining from Reference Videos}
\label{ssec:knowledge-mining-from-video}
Reference videos contain rich contextual knowledge essential for PID,
capturing users' identities, candidate personal items,
and user-item interactions essential to infer ownership.
Rather than using classic computer vision techniques,
we leverage an MLLM to extract this structured knowledge.

\begin{wrapfigure}{r}{0.580\linewidth}
  \centering
\vspace{-12mm}
  \includegraphics[width=\linewidth]{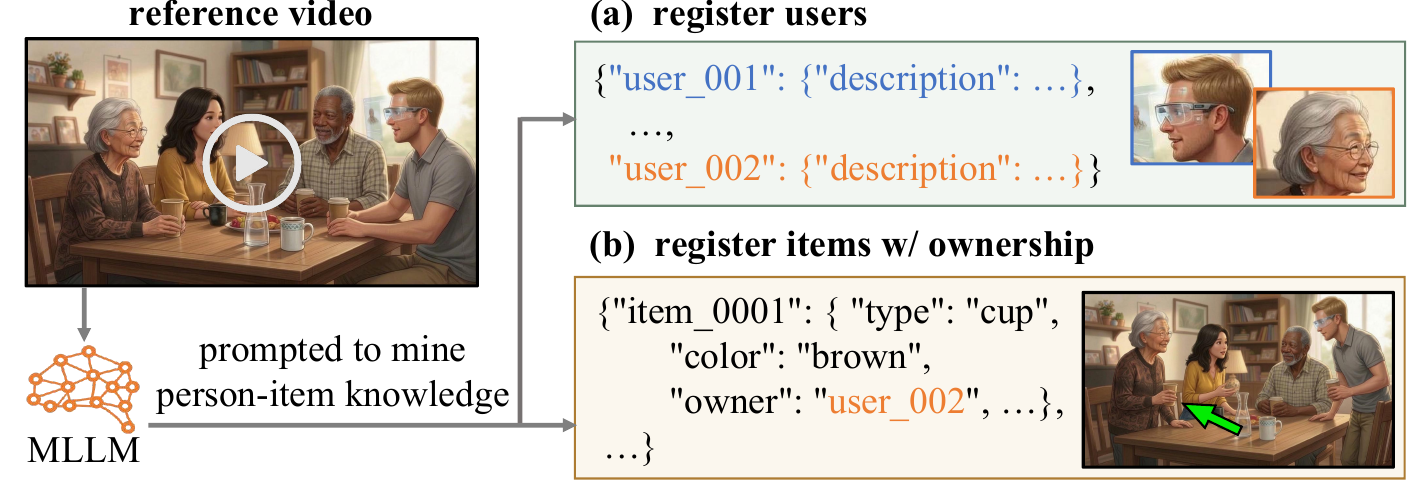}
\vspace{-9mm}
  \caption{\small 
  \textbf{Person-item knowledge mining and storage.}
  We leverage an MLLM to analyze reference videos to mine person-item knowledge, including users' images and descriptions, and 
  item descriptions, images and ownership.
  We structure an external memory bank in JSON format to store this knowledge. The memory bank is used later to detect personal items at test time.
  }
  \label{fig:pid_ai_memory_storage}
\vspace{-0mm}
\end{wrapfigure}

Specifically, for explicit memory under UAR (Fig.~\ref{fig:pid_ai_memory_storage}), 
we prompt the MLLM to identify distinct individuals,
generate descriptions of their appearance, and crop their corresponding visual regions from video frames,
thereby automatically registering the users observed in the videos.
Conditioned on these registered users, 
we next prompt the MLLM to detect all items involved in user interactions, reason about item ownership, and describe each item's visual appearance.
Further, to ground its ownership reasoning,
the MLLM is prompted to select the most representative video frame as visual evidence.
The resulting descriptions and visual crops form the external repository of explicit memory; implicit memory instead retains the video-understanding dialogue context, as described next.
Refer to Suppl.~\S\ref{supplsec:method_details} for further implementation details.

\subsection{Design of Memory Bank}
\label{ssec:memory-design}

\textbf{Explicit, human-readable memory (ExplicitMem).} 
We store the mined knowledge as structured textual descriptions and visual crops within JSON files (\cref{fig:pid_ai_memory_storage}-right).
Notably,
modern MLLMs can be readily prompted to generate JSON outputs, eliminating the need for manual encoding.
During inference,
the MLLM leverages these stored JSON files to identify target users and retrieve their personal items, based on the FUN queries of specific users.

\textbf{Implicit, KV-cache memory (ImplicitMem).} 
We retain the video-understanding dialogue context, including the sampled frames and generated descriptions, as implicit memory for subsequent queries.
The MLLM reasons over this context during localization, while KV caching permits reuse of matching attention states \cite{zhang2023h2o, shi2024keep, brandon2024reducing, xiao2024efficient, li2024snapkv, liu2024cachegen, liu2024deepseek}.
Unlike ExplicitMem, this design does not require a separately structured JSON memory.

\subsection{Memory-based Inference}
\label{ssec:final-inferenec}

\begin{wrapfigure}{r}{0.580\linewidth}
  \centering
    \vspace{-4mm}
  \includegraphics[width=\linewidth]{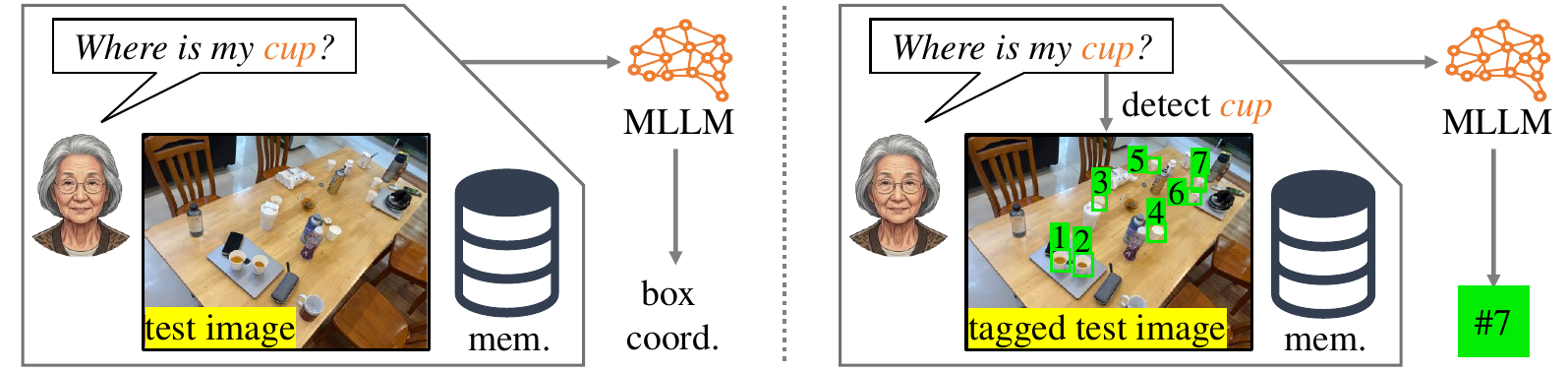}
    \vspace{-7mm}
  \caption{\small \textbf{Methods for memory-based inference.}
  We compare direct box-coordinate prediction with tagged proposal selection.
  As is well known,
  MLLMs struggle with spatial reasoning, particularly direct prediction of object locations \cite{sun2024alpha, ma2025auto};
  instead, selecting tagged regions turns out to be a better method (\Cref{tab:pidove-main}).
  }
  \label{fig:pid_ai_inference}
\vspace{-1mm}
\end{wrapfigure}

With an external memory bank storing PID-relevant knowledge,
we introduce two MLLM-based inference strategies to localize target instances from users' FUN queries:
\begin{enumerate}[leftmargin=12pt, topsep=1pt, itemsep=2pt, parsep=0pt]
    \item 
    \textbf{Box coordinate prediction (Box)}, 
    shown in Fig.~\ref{fig:pid_ai_inference}-left,  
    prompts the MLLM to directly output the bounding box coordinates w.r.t the user's profile image $u$,
    their FUN query  $q$, the test image $I$, and
    the memory bank.

    \item \textbf{Tagged proposal selection (Tag)}, shown in Fig.~\ref{fig:pid_ai_inference}-right, parses a FUN query $q$ to identify the target object category and applies a detector to detect all candidate instances of this category in the test image $I$. It then overlays tags onto these candidate boxes and uses the tagged image, alongside the user's profile image $u$ and the memory bank, to prompt the MLLM to select the relevant tags and retrieve their bounding boxes as the PID output.
\end{enumerate}

\section{Experiments}
\label{sec:pidove-experiments}
To gain deeper insights into the PID problem,
we evaluate our developed baselines and methods.
The results reveal the limitations of these methods and future research directions.

\textbf{Important details in implementation.}
Unless otherwise specified, all MLLM-based methods use the MLLM Qwen3.8-27B with tensor parallelism across two NVIDIA RTX A6000 GPUs (each with 48GB memory). 
For tag-based methods, Qwen3.5-0.8B~\cite{qwen3.5} extracts a target-object phrase from the FUN query, which serves as the text prompt for
LocateAnything~\cite{wang2026locateanything} to generate candidate boxes.

\begin{table}[t]
\centering
\small
\setlength{\tabcolsep}{0.9em}
\caption{\small
\textbf{PID results (mAP) on the VOPID test set}.
All methods are tuned on the VOPID validation set and 
use  Qwen3.8-27B as the underlying MLLM (except the classic baseline).
Both baselines apply across the UPR and UAR settings;
highest numbers within each diagnostic subset and setting are marked in bold.
Key results include:
(1) the modern baseline that prompts Qwen3.8 directly to output PID results yields surprisingly poor results, significantly underperforming the classic baseline;
(2) direct box prediction (Box) significantly underperforms selecting tagged proposals (Tag);
(3) with tagged proposal selection, KV-cache memory (ImplicitMem) outperforms structured human-readable memory (ExplicitMem);
(4) the results do not show which setting is harder between UAR and UPR.
Lastly, we report two reference measures:
(a) the proposal recall upper bound of LocateAnything used in ExplicitMem-Tag and ImplicitMem-Tag,
which assumes oracle matching and highlights significant headroom for improvement;
and (b) human performance, 
demonstrating both the challenge of PID and the quality of the VOPID benchmark (see the main text for details).
}
\label{tab:pidove-main}

\begin{tabular}{@{}llrrrrrrrr@{}}
\toprule

& & \textbf{overall} & \multicolumn{4}{c}{\textbf{target size and absence}} & \multicolumn{3}{c}{\textbf{occlusion level}} \\

\cmidrule(lr){3-3}\cmidrule(lr){4-7}\cmidrule(l){8-10}
& method & all & small & medium & large & absent & none & partial & heavy \\
\midrule

\multirow{2}{*}{\rotatebox{90}{\makecell{\textbf{base}\\\textbf{line}}}} 
& classic baseline & 24.47 & 17.98 & 21.79 & 38.40 & \textbf{64.71} & 22.09 & 20.80 & 7.93 \\
& modern baseline  & 4.04  & 1.42  & 5.54  & 25.60 & 0.00  & 3.85  & 6.67  & 2.41 \\
\midrule

\multirow{4}{*}{\rotatebox{90}{\makecell{\textbf{UPR}\\\textbf{setting}}}} 
& ExplicitMem-Box & 8.77  & 3.81  & 13.66 & 30.00 & 8.82  & 9.54  & 8.85  & 2.07 \\
& ImplicitMem-Box & 5.45  & 1.33  & 5.18  & 34.00 & 11.76 & 4.52  & 6.44  & 2.76 \\
& ExplicitMem-Tag  & 28.28 & 23.30 & 34.55 & 50.00 & \textbf{23.53} & 30.96 & 31.03 & 3.45 \\
& ImplicitMem-Tag  & \textbf{30.33} & \textbf{25.14} & \textbf{36.34} & \textbf{58.00} & \textbf{23.53} & \textbf{31.59} & \textbf{33.33} & \textbf{18.97} \\
\midrule

\multirow{4}{*}{\rotatebox{90}{\makecell{\textbf{UAR}\\\textbf{setting}}}} 
& ExplicitMem-Box & 8.07  & 3.85  & 13.21 & 28.80 & 2.94  & 8.33  & 10.00 & 6.21 \\
& ImplicitMem-Box & 4.99  & 1.38  & 4.73  & 28.40 & 11.76 & 3.68  & 6.67  & 2.76 \\
& ExplicitMem-Tag  & 27.76 & 21.51 & 35.45 & 53.60 & 23.53 & 29.83 & 30.80 & 6.55 \\
& ImplicitMem-Tag  & \textbf{31.23} & \textbf{24.50} & \textbf{37.14} & \textbf{70.00} & \textbf{26.47} & \textbf{33.51} & \textbf{30.92} & \textbf{18.97} \\
\midrule

\multirow{3}{*}{\rotatebox{90}{\makecell{\textbf{\textcolor{gray}{ref.}}\\\textbf{\textcolor{gray}{perf.}}}}} & \textcolor{gray}{recall} & \textcolor{gray}{60.28} & \textcolor{gray}{51.79} & \textcolor{gray}{60.09} & \textcolor{gray}{81.20} & \textcolor{gray}{n/a} & \textcolor{gray}{60.79} & \textcolor{gray}{53.56} & \textcolor{gray}{29.66} \\

& \textcolor{gray}{closed-book human} & \textcolor{gray}{82.91} & \textcolor{gray}{82.98} & \textcolor{gray}{97.83} & \textcolor{gray}{100.00} & \textcolor{gray}{48.28} & \textcolor{gray}{90.22} & \textcolor{gray}{92.10} & \textcolor{gray}{54.21} \\

& \textcolor{gray}{open-book human} & \textcolor{gray}{91.76} & \textcolor{gray}{94.66} & \textcolor{gray}{99.83} & \textcolor{gray}{100.00} & \textcolor{gray}{55.56} & \textcolor{gray}{97.42} & \textcolor{gray}{97.98} & \textcolor{gray}{79.64} \\

\bottomrule
\end{tabular}
\vspace{-0mm}
\end{table}

\subsection{Results}
\label{sec:pidove-benchmark}
\begin{figure}[t]
\centering
\includegraphics[width=\textwidth]{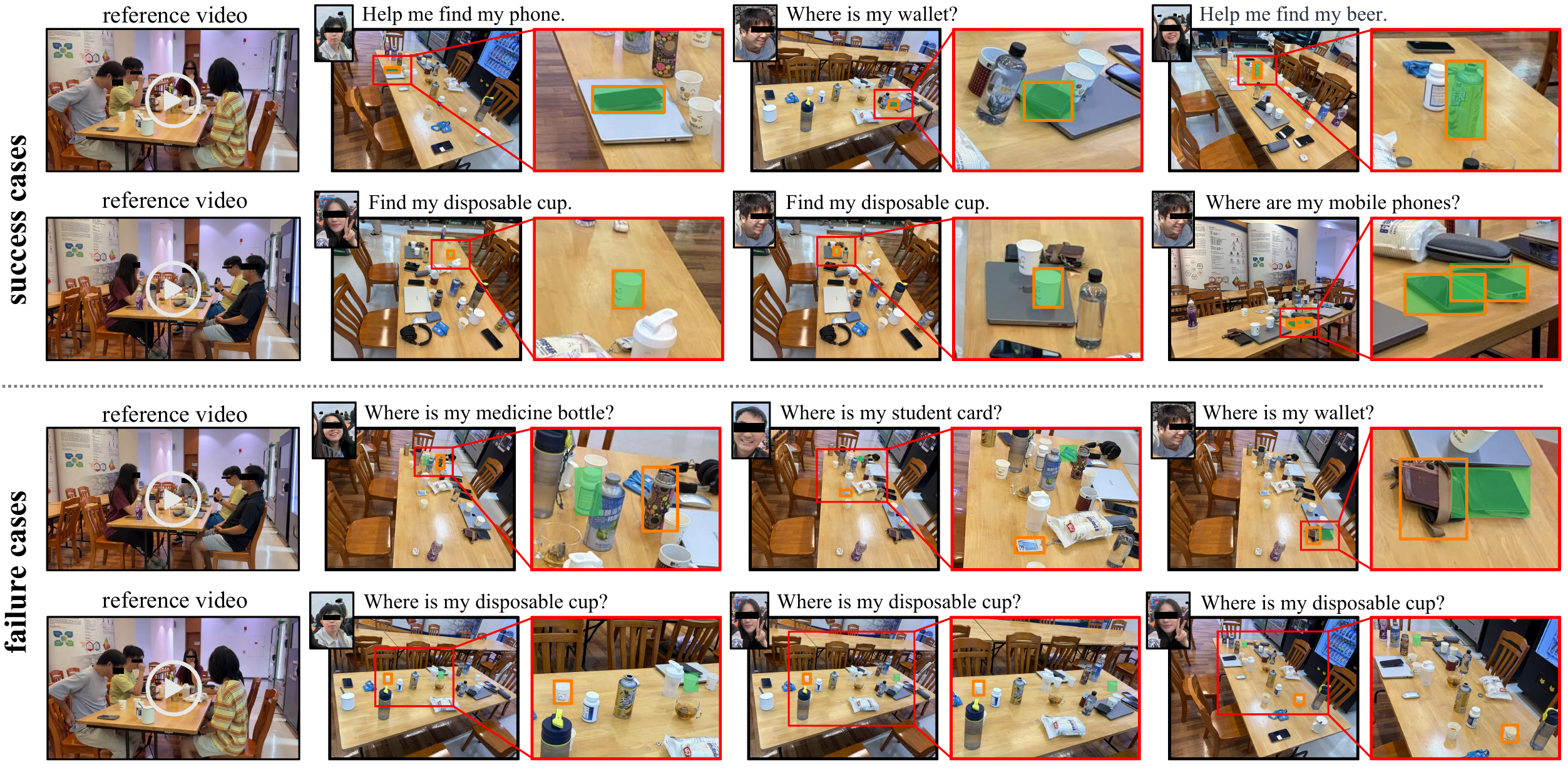}
\vspace{-9mm}
\caption{\small
\textbf{Qualitative results of ImplicitMem-Tag} under the UPR setting.
We show some success and failure cases in the top and bottom panels:
each row displays a reference video and three FUN queries with the corresponding users' profiles.
Ground-truth boxes and detections are marked by \textcolor{green!70!black}{green masks} and \textcolor{orange}{orange boxes}, respectively.
}
\label{fig:pid-qualitative-results}
\vspace{-0mm}
\end{figure}

Table~\ref{tab:pidove-main} compares our developed baselines and methods, with breakdown results w.r.t diagnostic subsets under both UPR and UAR settings.
Consistent with intuition,
smaller and heavily occluded instances are generally more difficult to detect across settings.
Yet, 
the classic baseline achieves better performance on instance-absent queries than other methods that leverage an MLLM, likely due to the well-known observation that MLLMs tend to output hallucinated results, i.e., fabricating detection boxes for non-existing instances to respond to such queries,
and lack the ability to say ``I don't know''.
\cref{fig:pid-qualitative-results} displays visual results on success and failure cases.
We present insightful analyses below.

\textbf{The modern baseline},
which prompts Qwen3.8 directly to output PID results,
yields rather poor results, significantly underperforming the classic baseline.
This implies a limitation of modern AI systems for solving complex tasks such as PID and suggests future research to empower these systems toward providing personalized services.

\textbf{Human performance.}
The final rows of Table~\ref{tab:pidove-main} report re-annotation accuracy of
(1) closed-book humans, who watched each reference video once before re-annotating test images for the corresponding FUN queries,
and (2) open-book humans, who were permitted to freely refer back to videos while re-annotating test images (Suppl.~\S\ref{supplsec:human} provides details on the evaluation).
The high accuracy of open-book humans validates the annotation quality of VOPID,
whereas the lower accuracy of closed-book humans underscores both the challenge and the value of PID.
Notably, both exhibit low accuracy on instance-absent queries,
implying a human insufficiency to handle nonexistent objects.

\textbf{Box selection vs. box prediction.}
Direct box prediction (e.g., ExplicitMem-Box) significantly underperforms the indirect approach that prompts the MLLM to select tagged boxes (e.g., ExplicitMem-Tag).
This reveals that MLLMs struggle to output precise coordinates of target objects' positions,
aligning with observations reported in the recent literature \cite{sun2024alpha, ma2025auto}.
Moreover, 
the third-to-last row in Table~\ref{tab:pidove-main} reports recall of LocateAnything as an upper-bound performance of the ``box selection'' approaches.
The gap confirms the pragmatic solution by leveraging a specialized computer vision method to empower MLLMs for precise spatial reasoning;
meanwhile, its far-from-perfect performance suggests future research improve not only MLLMs' capability of precise spatial predictions but also foundation detectors to detect target instances inferred from FUN queries.

\textbf{KV-cache  memory vs. human-readable memory.}
With the former,
ImplicitMem-Tag outperforms ExplicitMem-Tag, while the latter offers independently reloadable textual and visual memory for practical deployment.
These Tag results suggest a trade-off between PID performance and memory inspectability, motivating future integration of both paradigms.

\textbf{Difficulty of UAR vs. UPR settings.}
Comparisons of each method across the two settings do not clearly indicate which setting is more challenging.
Notably,
while methods under UAR detect individuals from reference videos and register them automatically,
the registry file (Fig.~\ref{fig:pid_ai_memory_storage}a) may contain repeated entries for the same person.
Yet, this does not affect person re-ID required at test time for matching users' profile images and stored crops of detected users.

\subsection{More Analyses}
\label{sec:pidove-ablation}
\textbf{Wall-clock time.}
Table~\ref{tab:pidove-efficiency} 
compares different methods w.r.t their wall-clock time at two stages: knowledge mining from videos (KMV) and inference.
The KMV stage takes a significant amount of time for methods to extract user-item knowledge from reference videos;
yet, this process can run in the background, in a lifelong manner,
without interrupting inference.
Regarding inference latency, 
except the classic baseline, which is significantly slower due to the need to process the entire reference video for each query,
other methods have an inference latency of 4-8s.
Reducing this latency will improve the user experience in the real world and can be achieved with optimized model deployment.

\textbf{Results of using different MLLMs.}
Table~\ref{tab:pidove-backbones} compares 
Qwen3.6 \cite{qwen36_35b_a3b} and Qwen3.8 \cite{qwen3.8} as the underlying MLLMs for different methods under the UPR and UAR settings.
The findings and conclusions, presented in \S\ref{sec:pidove-benchmark}, still hold.

\begin{table}[t]
    \centering
    \begin{minipage}[c]{0.41\linewidth}
        \centering
        \small
        \setlength{\tabcolsep}{1.2mm}
        \renewcommand{\arraystretch}{1.0}
        \caption{\small \textbf{Efficiency under UPR.} KMV: video knowledge mining (s/video); inference: query processing (s/query), including video processing for baselines. For Tag methods, inference includes Qwen3.5-0.8B phrase extraction, LocateAnything proposal generation, tag rendering, and MLLM selection. Performance: mAP.}
        \label{tab:pidove-efficiency}
        \resizebox{\linewidth}{!}{%
        \begin{tabular}{lrrr}
            \toprule
            Method & KMV $\downarrow$ & inference $\downarrow$ & mAP $\uparrow$ \\
            \midrule
            classic baseline & n/a & 87.37 & 24.47 \\
            modern baseline & n/a & 7.68 & 4.04 \\
            \midrule
            ExplicitMem-Box & 134.68 & 4.60 & 8.77 \\
            ImplicitMem-Box & 66.56 & 4.35 & 5.45 \\
            ExplicitMem-Tag & 132.48 & 6.74 & 28.28 \\
            ImplicitMem-Tag & 66.55 & 6.14 & 30.33 \\
            \bottomrule
        \end{tabular}%
        }
    \end{minipage}\hfill
    \begin{minipage}[c]{0.56\linewidth}
        \centering
        \small
        \setlength{\tabcolsep}{1.5mm}
        \renewcommand{\arraystretch}{1.0}
        \caption{\small \textbf{PID results (mAP) across different MLLMs on the VOPID test set}. We compare two open-source MLLMs: Qwen3.6-35B-A3B \cite{qwen36_35b_a3b} and Qwen3.8-27B \cite{qwen3.8}, under both the UPR and UAR settings. Unlike the modern baseline, which prompts the MLLM to directly output PID predictions, our modular methods utilize the MLLM to parse FUN queries, mine user-item knowledge from reference videos, and match user profile images against the memory bank during inference. The salient findings from Table~\ref{tab:pidove-main} still hold across backbones, with the more recent Qwen3.8 outperforming Qwen3.6 within our modular pipeline.}
        \label{tab:pidove-backbones}
        \resizebox{\linewidth}{!}{%
        \begin{tabular}{lrrrr}
            \toprule
            \multirow{2}{*}{Method}
            & \multicolumn{2}{c}{\textbf{UPR setting}}
            & \multicolumn{2}{c}{\textbf{UAR setting}} \\
            \cmidrule(lr){2-3}\cmidrule(l){4-5}
            & Qwen3.6 & Qwen3.8 & Qwen3.6 & Qwen3.8 \\
            \midrule
            modern baseline & 4.32 & 4.04 & 4.32 & 4.04 \\
            \midrule
            ExplicitMem-Box & 5.55 & 8.77 & 4.81 & 8.07 \\
            ImplicitMem-Box & 3.50 & 5.45 & 4.01 & 4.99 \\
            ExplicitMem-Tag & 22.26 & 28.28 & 22.93 & 27.76 \\
            ImplicitMem-Tag & 27.07 & 30.33 & 26.17 & 31.23 \\
            \bottomrule
        \end{tabular}%
        }
    \end{minipage}
    \vspace{-0mm}
\end{table}

\section{Impacts, Limitations and Future Work}
\textbf{Societal impacts.} 
The formulated PID problem helps lay the foundation for upcoming AI systems capable of delivering individualized, privacy-sensitive services.
Advancing research in PID will help democratize personalized AI technologies and meaningfully enhance everyday quality of life.
Furthermore, our VOPID benchmark provides a valuable resource for evaluating future AI products on precise spatial reasoning and perception in complex, personalized scenarios.
Regarding potential negative impacts,
PID methods process privacy-sensitive, personally identifiable information;
even when executed offline on local devices,
network connectivity could expose stored memory banks to data leakage under malicious attacks.
Moreover, 
our methods extensively leverage pretrained MLLMs,
which may carry inherent social and demographic biases from their pretraining data,
potentially surfacing as biased appearance descriptions in intermediate pipeline outputs.

\textbf{Limitations and future work.}
We outline several limitations in our dataset construction and methodology that point to promising directions for future research.
First,
constructing the VOPID dataset involves research-related simplifications that constrain real-world generalization:
(i) we curate short clips dense with social interactions from longer recordings rather than evaluating on continuous, unedited video streams;
(ii) we condition test-time FUN queries solely on user profile images, omitting acoustic cues (e.g., speaker voice) that provide valuable identity signals;
and (iii) all videos are captured using a single smartphone model rather than heterogeneous camera devices.
Future benchmarks can relax these constraints to better reflect unconstrained deployment.
Second,
while our formulation focuses exclusively on individual belongings,
everyday environments also contain communal objects, such as a shared TV remote control;
extending PID to jointly model personal and shared items is a natural next step.
Third,
we impose no storage budget on the external memory bank;
future work could investigate memory-bounded, life-long settings,
e.g., capping memory at 32GB, which remains ample for tens of thousands of items alongside compression strategies.
Fourth,
our empirical results reveal clear complementary trade-offs:
the classic baseline excels at rejecting instance-absent queries compared to MLLM-based methods,
yet lags substantially on present-item queries.
Similarly,
while a KV-cache memory (e.g., adopted by ImplicitMem-Tag) and structured JSON memory (e.g., adopted by ExplicitMem-Tag) offer distinct advantages in accuracy versus human interpretability,
hybridizing these paradigms presents a compelling direction to achieve both high PID accuracy and transparent, inspectable memory.

\section{Conclusion}
We formulate the novel problem Personal Item Detection (PID),
motivated to help individuals locate personal belongings based on their fuzzy, user-friendly, natural-language (FUN) queries, such as ``where is my wallet?'' asked by Sophie, who is just one of multiple users. 
To support PID research,
we introduce the challenging benchmark VOPID,
featuring multi-person and multi-object interaction, and FUN queries.
We develop first-principles baselines using both classic computer vision techniques and modern AI agents, and
present a modular pipeline where we study improved modules using foundation models and MLLMs.
Comprehensive results reveal limitations of these methods and outline promising research directions.

\section*{Acknowledgments}
SK gratefully acknowledges support from 
the CK Foundation and the Plato Initiative.
This work is supported by 
the Science and Technology Development Fund of Macau (0058/2025/RIA2, 0067/2024/ITP2),  and
the University of Macau (MYRG-CRG2026-00038-FIC, MYRG-GRG2026-00236-FIC, SRG2023-00044-FST).

\section*{LLM Usage Statement}
This statement discloses the use of Large Language Models (LLMs) and other generative AI tools, collectively referred to as ``AI tools'' hereafter.

Our work includes a study on the performance of modern AI tools, in which we analyze the performance of Qwen3.5-0.8B, Qwen3.6-35B-A3B, and Qwen3.8-27B.
Moreover, for better visual illustrations without personally identifiable information in figures, we use LLM-generated photos rather than real photos.
Other than these, we used Gemini strictly for language refinement and grammar corrections.
We thoroughly reviewed all AI-assisted revisions to ensure they precisely reflected the original concepts, methodology descriptions, and conclusions of this work.
AI tools were not used to generate research ideas, hypotheses, data, experimental designs, tables, code, or any core technical contributions.

{
    \small
    \bibliographystyle{plainnat}
    \bibliography{arxiv}
}

\clearpage
\setcounter{page}{1}
\renewcommand{\thesection}{\Alph{section}}
\renewcommand{\theHsection}{\Alph{section}}
\setcounter{section}{0}

{
\begin{center}
\Large
\textbf{Appendix}
\end{center}
}

The appendix supplements our main paper with more details. 
It is organized as follows:
\begin{itemize}
    \item Section~\ref{supplsec:datasets}: dataset construction and statistics.
    \item Section~\ref{supplsec:annotation}: annotation guidelines and ownership evidence.
    \item Section~\ref{supplsec:evaluation}: evaluation protocol and mAP definition.
    \item Section~\ref{supplsec:method_details}: implementation details and prompt templates.
    \item Section~\ref{supplsec:more_results}: additional results, controlled ablations, and visualizations.
    \item Section~\ref{supplsec:human}: human evaluation.
    \item Section~\ref{supplsec:code}: code availability and experimental environment.
\end{itemize}

\section{Dataset Construction and Statistics}
\label{supplsec:datasets}
\subsection{Collection and Query Construction}

\textbf{Reference videos.} The VOPID dataset comprises 41 reference videos and 519 FUN queries collected from 9 participants.
We recorded daily interactions in living rooms, shared offices, recreation lounges, and meeting rooms using a smartphone mounted on a high tripod.
We selected these videos from longer raw recordings to capture informative person-object and person-person interactions.

\textbf{Test images and user queries.} After each video recording, we captured the scenes with the same device from different viewpoints and distances, introducing changes in object appearance and image coordinates between reference videos and test images. Based on these test images, participants identified their belongings and formulated the corresponding questions. Consequently, each test sample consists of a user profile image, a FUN query, and a test image.

To reflect real-world deployment settings, systems receive reference videos without ownership annotations. 
A single reference video may show different users using items of the same category.

\begin{wraptable}{r}{0.55\linewidth}
\centering\small
\setlength{\tabcolsep}{3pt}
\renewcommand{\arraystretch}{1.05}
\vspace{-15mm}
\caption{\small \textbf{VOPID dataset statistics.} Val.: validation split. Box annotations: total number of target bounding-box instances across all queries. $\dagger$: number of reference videos involving same-category items from different users, along with their corresponding queries with targets present (each query counted once).}
\label{tab:supp-dataset-statistics}
\begin{tabular*}{\linewidth}{@{\extracolsep{\fill}}lrr@{}}
\toprule
Statistic & Val. & Test \\
\midrule
Reference videos & 10 & 31 \\
Queries & 130 & 389 \\
\quad Single-target & 106 & 319 \\
\quad Multi-target & 12 & 36 \\
\quad Empty-target & 12 & 34 \\
Box annotations & 130 & 394 \\
\midrule
\multicolumn{3}{@{}l}{\textit{Same-category items across users}$^\dagger$} \\
\quad Reference videos & 7 & 23 \\
\quad Queries & 67 & 216 \\
\midrule
Minimum clip duration (s) & 27.87 & 20.63 \\
Maximum clip duration (s) & 186.07 & 217.07 \\
\bottomrule
\end{tabular*}
\vspace{-3mm}
\end{wraptable}

\subsection{Splits and Statistics}
We divide VOPID into a validation set of 10 videos (130 queries) and a test set of 31 videos (389 queries), with detailed split statistics provided in Table~\ref{tab:supp-dataset-statistics}.
The two splits contain distinct reference videos.
All prompts and hyperparameters are selected on the validation set and fixed before test evaluation.

A distinctive feature of VOPID is that a single reference video often contains same-category personal items belonging to different users.
Queries about these items require systems to infer ownership from observed interactions and distinguish the designated user's belongings from same-category distractors.
Such cases span 7 validation videos and 23 test videos, accounting for 67 and 216 queries, respectively.

\subsection{Diagnostic Subsets}
\label{supplsec:diagnostics}
We analyze test performance across target scale, occlusion, and absence.
For a ground-truth bounding box $(x_1, y_1, x_2, y_2)$ in an image of width $W$ and height $H$, its relative area is defined as
\begin{equation}
r=\frac{(x_2-x_1)(y_2-y_1)}{WH},
\end{equation}

where $(x_1, y_1)$ and $(x_2, y_2)$ denote the top-left and bottom-right corner coordinates, respectively. Based on $r$, target-present queries are partitioned into small ($r$ $<$ 0.01; 218 queries), medium (0.01 $\leq$ $r$ $<$ 0.05; 112 queries), and large ($r$ $\geq$ 0.05; 25 queries) subsets, with multi-target queries assigned according to their smallest target.

Occlusion subsets are defined using the manually annotated occluded fraction $o \in [0, 1]$: unoccluded ($o$ $=$ 0; 239 queries), partially occluded (0 $<$ $o$ $<$ 0.5; 87 queries), and heavily occluded ($o$ $\geq$ 0.5; 29 queries), where multi-target queries take the maximum occluded fraction across targets. Finally, the 34 empty-target queries are evaluated separately and excluded from both the scale and occlusion subsets, while all 389 test queries contribute to the overall score.

\begin{wraptable}{r}{0.55\linewidth}
\vspace{-5mm}
  \centering
  \includegraphics[width=\linewidth]{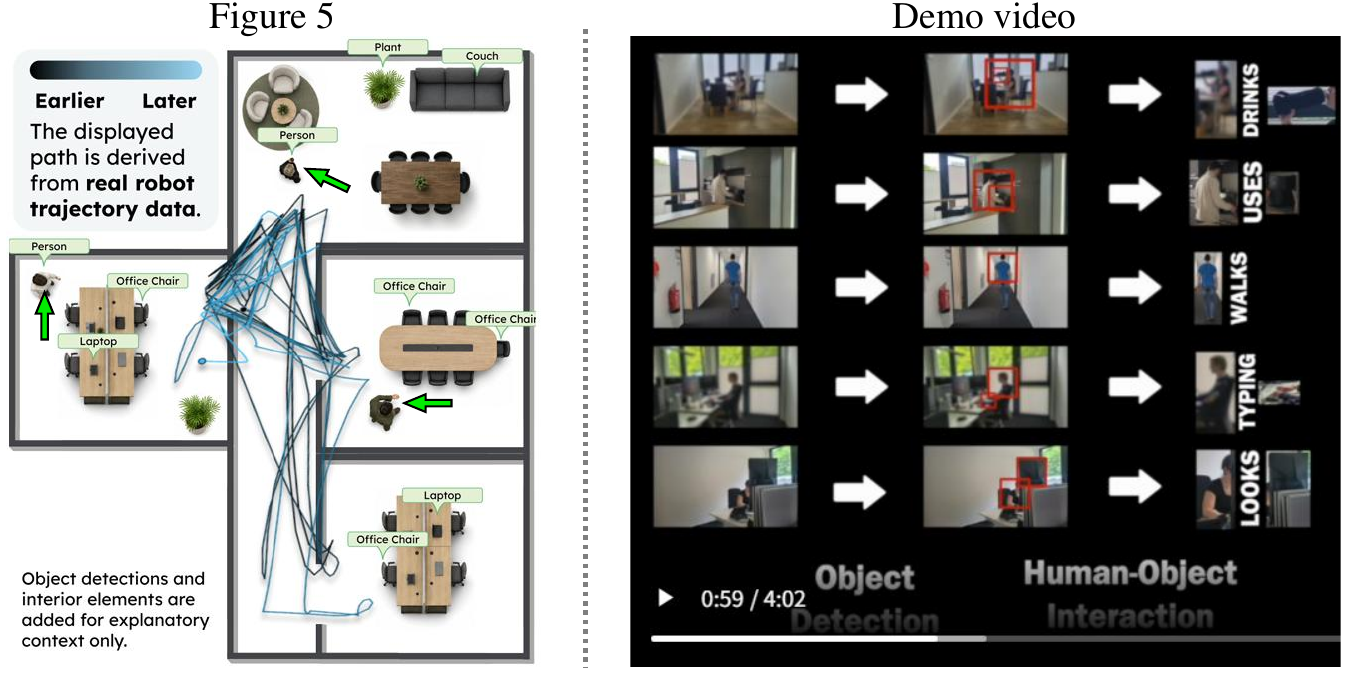}
\vspace{-9mm}
  \caption{\small \textbf{Real-world evaluation and visual examples in COOL \cite{huber2026cool}.}  Left: Robot trajectory over a 4.5-hour deployment in a single office environment (adapted from Fig.~5 of COOL). Right: Representative frames from the project demo video (\url{https://samirahuber.github.io/cool/}), showing single-user interactions without clear multi-user dynamics.}
  \label{fig:compare_cool}
\vspace{-3mm}
\end{wraptable}

\subsection{Comparison with Concurrent Work}
\label{supplsec:comparison}
Although COOL \cite{huber2026cool} is framed around multi-person environments, its reported real-world evaluation and project demo video reflect a more constrained setting. Specifically, as illustrated in Fig.~\ref{fig:compare_cool}, the real-world deployment is confined to a single office environment that contains only a small number of people and objects. Across the reported real-world tasks and demo frames, visible interactions mainly depict single-person or separated interactions rather than concurrent multi-user dynamics and complex social interactions characteristic of unconstrained real-world scenarios. Moreover, COOL's raw real-world video recordings are not publicly released, whereas VOPID is publicly available to support reproducible evaluation under multi-user interactions.

\section{Annotation Guidelines}
\label{supplsec:annotation_guidlines}
\label{supplsec:annotation}
\subsection{Ownership and Query Interpretation}
Participants identify their own belongings during annotation. When multiple participants own items of the same category and more than one appears in a test image, the answer to a query such as ``Where is my cup?'' depends on the requester, and only the requester's item is annotated as a target. For plural queries such as ``Where are my cups?'', annotators draw a separate bounding box around each identifiable item belonging to the requester.

At test time, methods infer ownership without access to participant ownership annotations. The prompts instruct the MLLM to look for behavioral evidence in the reference video, such as a person drinking from a cup, and to leave ownership uncertain when evidence is insufficient. A person standing just near an item does not establish ownership, whereas the classic baseline uses spatial proximity as a heuristic to estimate ownership.

\subsection{Visibility and Modal Bounding Boxes}
\begin{figure}[t]
  \centering
  \includegraphics[width=0.94\linewidth]{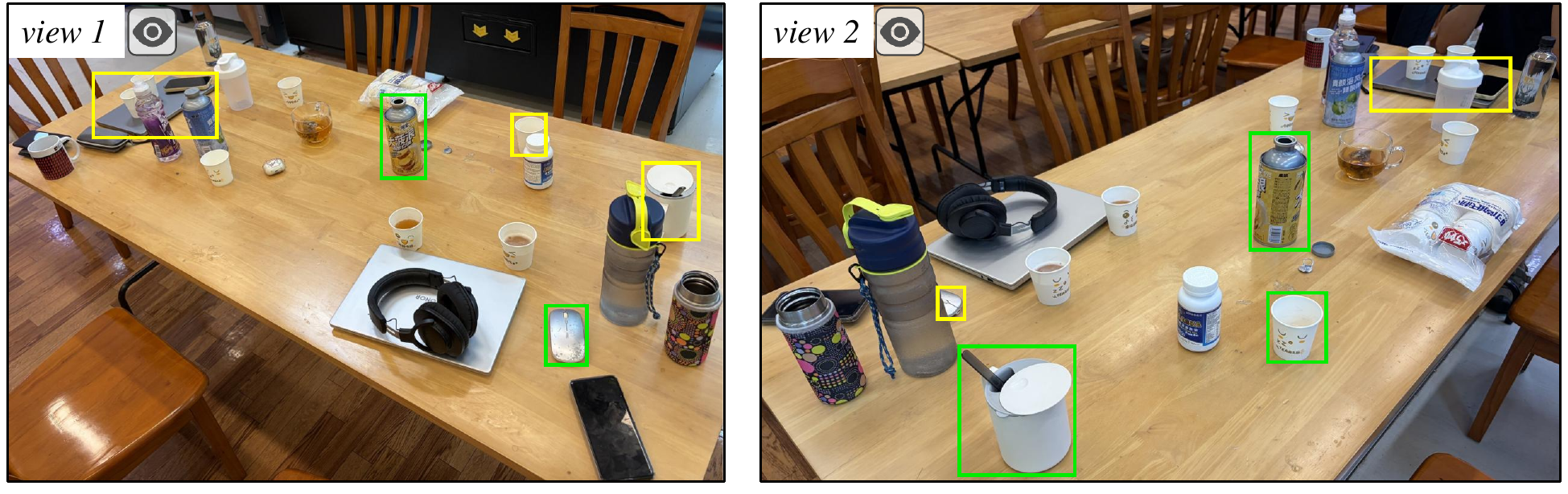}
  \vspace{-5mm}
  \caption{\small \textbf{Visual examples of our multi-view annotation guidelines.} \textcolor{green!70!black}{Green boxes} denote \textit{fully visible targets} enclosed by tight bounding boxes. \textcolor{yellow!80!black}{Yellow boxes} indicate \textit{partially visible targets} annotated via modal bounding boxes.
  }
  \label{fig:guidlines}
  \vspace{-0mm}
\end{figure}

We annotate each identifiable target instance in the test image with a tight bounding box around its visible extent (i.e., modal annotation), as illustrated in Fig.~\ref{fig:guidlines}. For a fully visible target, the box encloses the entire object. For a partially occluded yet identifiable target, a single box encloses all visible parts of that instance, even when an occluder separates them into multiple parts.

A query receives an empty annotation when none of the requested instances is identifiable in the test image.
This covers both targets absent from the current view and those that are completely occluded.
We assign the label ``absent'' to such queries in the diagnostic breakdown; consequently, a target may appear in the reference video or another view and still be marked ``absent'' in a given test image.

\subsection{Bounding-Box Representation}
For a query $q$ with $K$ identifiable target instances in the test image, the ground-truth annotation is represented as the set $G_q=\{g_1,\ldots,g_K\}$, with one box per identifiable instance. For an empty-target query, $G_q=\varnothing$ (or an empty list in serialized outputs). Each box $g_i=[x_1,y_1,x_2,y_2]$ specifies the top-left and bottom-right corners in the pixel coordinates of the original test image.

\section{Evaluation Protocol}
\label{supplsec:evaluation}
\subsection{Strict Matching and mAP}
\label{supplsec:strict-metric}
VOPID uses the same ten IoU thresholds as COCO, but evaluates set-level query accuracy instead of confidence-ranked average precision. At each threshold, we measure the fraction of queries for which the predicted set exactly matches the ground-truth set, and average these fractions across all thresholds to report $\mathrm{mAP}$ (\%).

Let $P_q$ and $G_q$ denote the predicted and ground-truth boxes for query $q$.
When $G_q \neq \varnothing$, a valid, complete prediction is correct at IoU threshold $\tau$ if and only if $\vert{}P_q\vert{} = \vert{}G_q\vert{}$ and $P_q$ matches $G_q$ one-to-one with every pair having $\mathrm{IoU} \geq \tau$.
Box order is immaterial, whereas any missed target or extra detection makes the entire query incorrect at that threshold.
When $G_q=\varnothing$, only a valid, explicit prediction of no boxes is correct. Invalid or incomplete outputs receive zero score at every threshold.

Let $c(q,\tau)$ equal one if query $q$ is correct at threshold $\tau$ and zero otherwise.
We report $\mathrm{mAP}$ (\%) by averaging this score over the queries in $\mathcal{Q}$ and the ten IoU thresholds $\mathcal{T}=\{0.50,0.55,\ldots,0.95\}$, scaled to percentages as
\begin{equation}
\mathrm{mAP}(\mathcal{Q}) = \frac{100}{|\mathcal{Q}|\,|\mathcal{T}|} \sum_{q\in\mathcal{Q}}\sum_{\tau\in\mathcal{T}} c(q,\tau).
\end{equation}

\subsection{Empty Predictions and Invalid Responses}
For empty-target queries where no requested instance is identifiable in the test image, a correct response must explicitly return an empty list in the designated JSON field: Box methods output \verb|{"bboxes": []}|, and Tag methods output \verb|{"selected_tags": []}|.
On the 34 empty-target test queries, $\mathrm{mAP}$ therefore equals the percentage of queries with a valid empty prediction, regardless of the IoU threshold.
These queries are included in the overall $\mathrm{mAP}$ and reported as a separate ``absent'' subset, outside the scale and occlusion subsets.

Any missing, truncated, or malformed response receives zero score at all thresholds, including on empty-target queries. The same zero-score rule applies to outputs containing duplicate boxes or tags, invalid bounding-box coordinates, or tag IDs outside the candidate proposal set.

\subsection{Upper Bound for Tag Selection}
\label{supplsec:oracle}
Tag methods can localize a target only if LocateAnything proposes a suitable box for it. To quantify this limit, we evaluate an oracle that uses the ground truth to select from the same fixed proposals. At each IoU threshold, a target-present query is correct if every annotated instance can be matched to a distinct proposal that meets the threshold. The oracle can leave all other proposals unselected. Its score thus gives an upper bound for methods that select tags from this pool. This bound does not apply to methods that predict bounding-box coordinates directly.

The oracle achieves 56.48 $\mathrm{mAP}$ on the 355 target-present test queries. For each of the 34 empty-target test queries, the oracle knows from the annotation that the correct selection is empty and returns no boxes, regardless of what LocateAnything proposes. This yields 100.00 on this subset. Including these queries gives 60.28 $\mathrm{mAP}$ on the full test set of 389 queries.

\section{Implementation Details and Prompts}
\label{supplsec:method_details}
We use the method names from the main paper: classic baseline, modern baseline, ExplicitMem-Box/Tag, and ImplicitMem-Box/Tag. For the two-stage methods, Stage~1 builds video-derived memory via knowledge mining from videos (KMV), and Stage~2 performs query-time inference. The modern baseline has no dedicated memory-construction stage.

\subsection{Shared Runtime and Preprocessing}
\label{supplsec:sampling}
\textbf{Hardware and inference setup.} The experimental setup comprises two NVIDIA RTX A6000 GPUs (48\,GB each), with PyTorch~2.10.0, vLLM~0.19.1, and Transformers~5.5.4. For Qwen3.8-27B~\cite{qwen3.8} and Qwen3.6-35B-A3B~\cite{qwen36_35b_a3b}, we use BF16 inference and greedy decoding with temperature~0, top-$p=1$, seed~0, and thinking disabled.
The main MLLMs (Qwen3.8-27B and Qwen3.6-35B-A3B) run with two-way tensor parallelism via vLLM using a 32,768-token context window and maximum output lengths of 8,192 tokens in Stage~1 and 2,048 tokens in Stage~2. For target-phrase extraction, Qwen3.5-0.8B~\cite{qwen3.5} runs on a single GPU via Hugging Face Transformers with a maximum output length of 96 tokens and a repetition penalty of 1.05.

\textbf{Uniform video frame sampling.} We sample frames uniformly from the first to the last frame of each reference video, targeting approximately 1\,fps with a default cap of 48 frames. For longer videos, the samples remain evenly spaced over the full clip. Given $F$ decoded frames at a native frame rate of $f$ frames per second, let $L=(F-1)/f$ denote the elapsed duration between the first and last frames. The sampled frame count $n$ and zero-based frame indices $k_i$ ($i = 0, \ldots, n-1$) are defined as
\begin{equation}
\begin{aligned}
n   &= \min\!\left(48,\max\!\left(2,\lceil L\rceil+1\right)\right), \\
k_i &= \operatorname{round}\!\left(\frac{i(F-1)}{n-1}\right).
\end{aligned}
\end{equation}

\textbf{Input resolution.} We preserve the original aspect ratio of all visual inputs and resize the longest side to 448 pixels for video frames, 224 pixels for user profile images, and 1,280 pixels for test images.

\subsection{Target-Phrase Extraction}
\label{supplsec:phrase-extraction}
To prepare target-category proposals for the classic baseline and Stage~2 of the Tag methods, Qwen3.5-0.8B~\cite{qwen3.5} parses each FUN query into a structured JSON object containing \texttt{category}, \texttt{attributes}, and \texttt{prompt}. Specifically, \texttt{category} and \texttt{attributes} extract the target object class and any explicitly stated visual modifiers, while \texttt{prompt} combines them into a non-empty noun phrase of at most six words. Validation filters out prompts containing possessives, command verbs, person or identity terms, or spatial prepositions, with up to two generation attempts allowed upon validation failure.
We pass \texttt{prompt} to LocateAnything~\cite{wang2026locateanything} to generate candidate boxes in the test image. We cache the extracted phrase and candidate boxes for reuse across methods.

\subsection{Classic Baseline}
\label{supplsec:classic_baselines}

The classic baseline follows a modular pipeline for detection, tracking, and re-identification built on task-specific vision models rather than MLLM memory representations, operating with its own video-sampling and tracking settings separate from the MLLM configurations. It directly reuses the shared Qwen3.5-0.8B prompts and LocateAnything proposals (refer to \S\ref{supplsec:phrase-extraction}) without invoking any MLLM during inference.

In the reference video, persons are detected with the person detector
and tracked across frames using ByteTrack~\cite{zhang2022bytetrack},
while the user is identified by matching their profile image against face features along these tracks via ArcFace~\cite{deng2019arcface}.
In parallel, LocateAnything~\cite{wang2026locateanything} detects and tracks objects of the target category across video frames, and DINOv2~\cite{oquab2023dinov2} extracts appearance features for association and matching.
Spatial proximity and motion cues associate the target object track with the identified user track to form an instance memory bank.
To localize the item in the test image, the classic baseline matches DINOv2 features from the instance memory bank against the fixed LocateAnything proposals. It returns the matching boxes when the identity, ownership, and appearance scores meet their respective thresholds.

When multi-output mode is enabled, the classic baseline can output zero, one, or up to two bounding boxes (defaulting to a maximum of two boxes). Because instance memory bank matching is designed for single-instance retrieval, multi-target outputs are capped at a small fixed box count and do not support general or unrestricted multi-target reasoning.

\subsection{Modern Baseline}
\label{supplsec:modern_baselines}

For each FUN query, the modern baseline provides the MLLM~\cite{qwen3.8} with the sampled reference frames, the user's profile image, the test image, and the full question in a single request. The model uses the observed interactions to associate the user with the requested items and outputs a list of bounding boxes covering all identifiable requested instances in the test image, with one box per instance. Coordinates are normalized to a $0$--$1000$ grid. For example, two requested instances are represented as \texttt{\{"bboxes":[[x1,y1,x2,y2],[x1,y1,\\x2,y2]]\}}, whereas an empty target set is represented as \texttt{\{"bboxes":[]\}}. Each query is processed independently.

\subsection{ExplicitMem}
\label{supplsec:explicit}
In Stage~1, ExplicitMem builds a structured JSON memory from the reference video, capping registered items at 12 with one to two \texttt{evidence\_events} per item.
Each item entry records the item's visual attributes, \texttt{candidate\_owners} (which may stay empty when ownership is uncertain), its last observed state and location, and timestamped evidence events from which supporting video frames are extracted.
Fig.~\ref{fig:supp-registry-upr} and~\ref{fig:supp-registry-uar} illustrate Stage~1 records generated under the UPR and UAR settings for the same electric kettle on the VOPID test set.

\textbf{User pre-registration (UPR) setting.}
Stage~1 receives the sampled reference video together with a labeled profile image for every registered user (both \texttt{P001} and \texttt{P002} in this example), grounding \texttt{people\_registry} directly to supplied identities.
Because external profile images already define each user, the video-based \texttt{face\_reference} is optional.
In Fig.~\ref{fig:supp-registry-upr}, the model lists \texttt{P002} as a candidate owner and links this user to the water-pouring event (\texttt{E001}), while leaving \texttt{person\_id} empty in the placement event (\texttt{E002}).

\textbf{User auto-registration (UAR) setting.}
Stage~1 receives only the sampled video without pre-registered profile images or user IDs, requiring the model to discover participants and assign video-local IDs (such as \texttt{P002}).
As shown in Fig.~\ref{fig:supp-registry-uar}, \texttt{face\_reference} becomes mandatory to ground each discovered identity: it specifies frame \texttt{F00000260} and a $0$--$1000$ normalized face bounding box (\texttt{[720, 0, 820, 150]}), from which a face crop is extracted to represent \texttt{P002}.
In contrast to the UPR excerpt, the UAR record attributes both pouring water (\texttt{E001}) and placing the kettle on the table (\texttt{E002}) to this local identity.

At inference time in Stage~2, the MLLM receives the structured registry, extracted event frames, test image, and FUN query alongside setting-specific visual inputs: under UPR, it takes the user's profile image together with all pre-registered profile images; under UAR, where no pre-registered profile images exist, it takes the user's profile image together with the extracted video face crops to localize the target instances.

\begingroup
\makeatletter
\def\@xobeysp{\leavevmode\nobreak\hskip\fontdimen2\font\relax\penalty\z@}
\let\vopidOriginalVerbatim\@verbatim
\def\@verbatim{\vopidOriginalVerbatim\rightskip=0pt plus 1fil\relax}
\makeatother
\begin{figure*}[tbp]
\centering
\begin{minipage}[t]{.485\textwidth}
\vspace{1pt}
{\small\sffamily\bfseries UPR registry}\par
\vspace{3pt}{\color{black!35}\hrule height .4pt}
{\footnotesize
\begin{verbatim}
{
  "people_registry": [{
    "person_id": "P002",
    "facial_description":
      "Young woman with shoulder-length black hair, fair skin, and an oval face.",
    "face_reference": {
      "frame_id": "F00000087",
      "bbox": [100, 100, 600, 800]
    }
  }],
  "items_registry": [{
    "item_id": "I001",
    "item_name": "Electric Kettle",
    "item_type": "kettle",
    "appearance":
      "Silver and black electric kettle",
    "final_location":
      "On the table, center-right",
    "candidate_owners": ["P002"],
    "last_user_interacted": "P002",
    "evidence_events": [{
      "event_id": "E001",
      "start_sec": 8.670072,
      "end_sec": 10.103969,
      "person_id": "P002",
      "action": "pours water"
    }, {
      "event_id": "E002",
      "start_sec": 67.87666,
      "end_sec": 67.87666,
      "person_id": "",
      "action": "placed on table"
    }]
  }]
}
\end{verbatim}
}
{\color{black!35}\hrule height .4pt}
\caption{\textbf{UPR output for the electric kettle.} The excerpt shows one of the two users whose portraits were supplied during memory construction. The second event leaves the person ID empty.}
\label{fig:supp-registry-upr}
\end{minipage}\hfill
\begin{minipage}[t]{.485\textwidth}
\vspace{1pt}
{\small\sffamily\bfseries UAR registry}\par
\vspace{3pt}{\color{black!35}\hrule height .4pt}
{\footnotesize
\begin{verbatim}
{
  "people_registry": [{
    "person_id": "P002",
    "facial_description":
      "Shoulder-length black hair, fair skin, wearing a white t-shirt.",
    "face_reference": {
      "frame_id": "F00000260",
      "bbox": [720, 0, 820, 150]
    }
  }],
  "items_registry": [{
    "item_id": "I001",
    "item_name": "Electric kettle",
    "item_type": "kettle",
    "appearance":
      "Silver and black electric kettle.",
    "final_location":
      "On the table, center-left.",
    "candidate_owners": ["P002"],
    "last_user_interacted": "P002",
    "evidence_events": [{
      "event_id": "E001",
      "start_sec": 7.219502,
      "end_sec": 13.005108,
      "person_id": "P002",
      "action": "pours water"
    }, {
      "event_id": "E002",
      "start_sec": 14.439005,
      "end_sec": 14.439005,
      "person_id": "P002",
      "action": "places on table"
    }]
  }]
}
\end{verbatim}
}
{\color{black!35}\hrule height .4pt}
\caption{\textbf{UAR output for the same kettle.} The frame and box in \texttt{face\_reference} specify the crop used to represent the video-local \texttt{P002}.}
\label{fig:supp-registry-uar}
\label{fig:supp-registry-schema}
\end{minipage}
\end{figure*}
\endgroup

\subsection{ImplicitMem: KV-Cache Memory}
\label{supplsec:implicit}
ImplicitMem uses a two-turn dialogue to carry information from video understanding into localization.
In the first turn, the model processes the sampled reference video to describe observed persons, their interactions with individual items, and the latest item locations, after which the first-turn exchange is frozen as a shared prefix.
Each FUN query is then evaluated in an independent second turn, receiving the user's profile image, the test image, and the FUN query while reusing only the first-turn context generated from the corresponding reference video. All previous queries from other users and prior localization responses are excluded from this context.

\textbf{User pre-registration (UPR) setting.}
In the first turn, the model receives labeled profile images for all registered users alongside the sampled reference video, binding observed actions and item states directly to registered identities in the shared prefix. In the second turn, the requester's profile image is matched against these pre-registered identities to resolve ownership and localize the target.

\textbf{User auto-registration (UAR) setting.}
Without pre-registered profile images or user IDs in the first turn, the model relies solely on the sampled reference video and describes participants by their visual appearance and interactions in natural language.
In the second turn, the model receives the requester's profile image and matches it against these appearance-grounded descriptions in the frozen prefix without extracting explicit face crops or structured identity records.

Because the first-turn prefix remains unchanged across queries on the same video, vLLM can reuse the cached attention keys and values during second-turn inference without recomputing the shared video context.
However, retaining KV states for a growing collection of videos can saturate GPU memory, and once cached blocks are evicted or cleared, the prefix must be recomputed.
By contrast, ExplicitMem serializes its structured JSON registry and extracted evidence frames to disk for on-demand loading.
Consequently, while ImplicitMem-Tag attains higher test $\mathrm{mAP}$ than ExplicitMem-Tag under both UPR and UAR, ExplicitMem is better suited to persistent cross-session storage and revisiting videos after long intervals.

\subsection{Box and Tag Interfaces}
\label{supplsec:localization}
The Box and Tag interfaces differ in how the model localizes the requested items in the test image.
Under the Box interface, the model directly predicts bounding boxes as integer coordinates $(\hat{x}_1, \hat{y}_1, \hat{x}_2, \hat{y}_2)$ on a normalized $0$--$1000$ grid.
We rescale these normalized coordinates to original pixel coordinates $(x_1, y_1, x_2, y_2)$ using the test image's width $W$ and height $H$:
\begin{equation}
\begin{aligned}
x_i &= \operatorname{round}\!\left(\frac{\hat{x}_i W}{1000}\right), \\
y_i &= \operatorname{round}\!\left(\frac{\hat{y}_i H}{1000}\right), \quad i \in \{1, 2\}.
\end{aligned}
\end{equation}

The Tag interface uses candidate boxes generated by LocateAnything from the target phrase extracted by Qwen3.5-0.8B (\S\ref{supplsec:phrase-extraction}).
We overlay each candidate box on the test image with a distinct numeric tag. Given this tagged image, the user's profile image, the video memory, and the full FUN query, the localization MLLM selects the matching tag indices (\texttt{selected\_tags}, or an empty list if none matches), and the corresponding proposal boxes form the final prediction.

\subsection{Prompt Templates}
\label{supplsec:prompts}
We give the message sequences used with Qwen3.8-27B. Each setting is organized by memory method and then by Box or Tag output. Box and Tag share the entire Stage~1 request within a memory method. Brackets marking inputs such as [requester portrait] or [full FUN query] indicate where the corresponding image, text, or schema is inserted. Brackets within JSON examples are literal. Repeated images are shown once with their accompanying text.

\begingroup
\newsavebox{\vopidRoleBox}
\newcommand{\vopidFullTitle}[1]{%
  \par\addvspace{8pt}\noindent\textbf{#1}\par\nobreak
}
\newenvironment{vopidRoleCard}[1]{%
  \par\addvspace{5pt}%
  \begin{lrbox}{\vopidRoleBox}%
  \begin{minipage}{\dimexpr\linewidth-12.6pt\relax}%
  \normalfont\small\raggedright
  \setlength{\parindent}{0pt}%
  \setlength{\parskip}{2pt}%
  \textbf{#1}\par\nobreak\vspace{2pt}%
}{%
  \end{minipage}\end{lrbox}%
  \begingroup\setlength{\fboxsep}{6pt}\setlength{\fboxrule}{.3pt}%
  \noindent\fcolorbox{black!25}{black!3}{\usebox{\vopidRoleBox}}\par
  \endgroup\addvspace{6pt}%
}

\subsubsection{User Pre-registration (UPR)}
\label{supplsec:prompt-upr}
Stage~1 receives labeled profile images for all users registered for the reference video. Stage~2 receives the current requester's profile image with each FUN query.

\vopidFullTitle{ExplicitMem-Box and ExplicitMem-Tag: shared Stage~1}
\begin{vopidRoleCard}{System}
You are a visual assistant for personal-item localization.\par
Identify people using stable facial evidence,\allowbreak{} not clothing. Track distinct physical
items through visible holding,\allowbreak{} wearing,\allowbreak{} use,\allowbreak{} touch,\allowbreak{} pickup and placement. Proximity
alone does not establish a person-item association. Keep same-category instances
separate. Preserve explicit attributes in the original question. Do not invent
identity,\allowbreak{} interaction or ownership evidence. Inspect the current scene rather than
assuming an item's last video location is unchanged. Return every requested visible
physical instance separately; do not include another person's items,\allowbreak{} duplicate
instances or unsupported extras. Return an explicit empty set when unsupported.\par
No thinking,\allowbreak{} markdown or text outside the requested JSON object.\par
There are two tasks:\allowbreak{} BUILD\_\allowbreak{}MEMORY before any localization question,\allowbreak{} and LOCALIZE.\par
For BUILD\_\allowbreak{}MEMORY output the requested registry schema; for LOCALIZE output only
the requested localization schema. Historical memory is predicted evidence,\allowbreak{} not GT.\par
IDENTITY CONDITION:\allowbreak{} During BUILD\_\allowbreak{}MEMORY,\allowbreak{} named reference portraits of the people in this video
are provided. Use exactly the supplied person IDs. Match video people to these
portraits by faces. Do not assign an uncertain interaction to a person; use an
empty person ID/list. face\_\allowbreak{}reference may be null because portraits are supplied.
\end{vopidRoleCard}
\begin{vopidRoleCard}{User}
Registered profile:\allowbreak{} [\allowbreak{}registered ID]\allowbreak{}\par
[\allowbreak{}registered portrait; repeat for each registered user]\allowbreak{}\par
Video [\allowbreak{}frame ID]\allowbreak{} at [\allowbreak{}timestamp]\allowbreak{} seconds.\par
[\allowbreak{}sampled frame; repeat in chronological order]\allowbreak{}\par
BUILD\_\allowbreak{}MEMORY. Observe the entire supplied timestamped video sequence.\par
Output a people registry and an items registry. Describe each person's stable face.\par
Register at most 12 visibly interacted item instances,\allowbreak{} merge repeated observations,\allowbreak{}
and preserve their discriminative appearance and final location relative to fixed
scene landmarks. Each item has one or two evidence events:\allowbreak{} an informative
interaction and,\allowbreak{} when observed,\allowbreak{} final placement. Use actual printed seconds from
the video start,\allowbreak{} unique event IDs,\allowbreak{} and only registered person IDs or an empty ID.\par
Keep string fields concise. Output exactly the supplied JSON schema.\par
Output JSON schema:\allowbreak{}\par
[\allowbreak{}registry schema for this setting and video]\allowbreak{}
\end{vopidRoleCard}

\vopidFullTitle{ExplicitMem-Box: Stage~2}
\begin{vopidRoleCard}{System}
Locate the physical items requested in the final query. Match the current requester using facial appearance. Use recorded interactions and visual evidence to distinguish same-category items. Memory may be wrong; check the current scene and correct conflicts. Localize all matching instances separately. Return an empty array if the requested items are absent or unsupported. Follow the query's JSON output format.
\end{vopidRoleCard}
\begin{vopidRoleCard}{User}
Historical person and item registry:\allowbreak{}\par
[\allowbreak{}Stage 1 JSON registry]\allowbreak{}\par
Reference face for video person [\allowbreak{}person ID]\allowbreak{}.\par
[\allowbreak{}registered portrait; repeat for each person]\allowbreak{}\par
Supporting video frame [\allowbreak{}frame ID]\allowbreak{} at [\allowbreak{}timestamp]\allowbreak{} seconds; links:\allowbreak{} [\allowbreak{}item and event references]\allowbreak{}\par
[\allowbreak{}event support frame; repeat for each supplied frame]\allowbreak{}\par
Current localization query. The following two images are Image A and Image B.\par
Image A:\allowbreak{} current requester.\par
[\allowbreak{}requester portrait]\allowbreak{}\par
Image B:\allowbreak{} current scene.\par
[\allowbreak{}test image]\allowbreak{}\par
LOCALIZE. Image A is the current requester's portrait. Image B is the
current scene. Match Image A to video people using faces and the supplied memory;
identify the requested instances through supported person-item interactions.\par
Reference face IDs and video frame IDs are not test-image proposal tags.\par
Use all relevant evidence and preserve the complete original question.\par
Original question:\allowbreak{} [\allowbreak{}full FUN query]\allowbreak{}
Image B has no proposal tags. Return separate tight boxes for ALL requested
instances,\allowbreak{} not a union box. Use integer coordinates [\allowbreak{}x1,\allowbreak{}y1,\allowbreak{}x2,\allowbreak{}y2]\allowbreak{} normalized to
0-1000 relative to Image B. Output exactly \{\allowbreak{}"bboxes":\allowbreak{} [\allowbreak{}[\allowbreak{}x1,\allowbreak{}y1,\allowbreak{}x2,\allowbreak{}y2]\allowbreak{},\allowbreak{} ...]\allowbreak{}\}\allowbreak{} or
\{\allowbreak{}"bboxes":\allowbreak{} [\allowbreak{}]\allowbreak{}\}\allowbreak{}. No additional fields.
\end{vopidRoleCard}

\vopidFullTitle{ExplicitMem-Tag: Stage~2}
\begin{vopidRoleCard}{System}
Locate the physical items requested in the final query. Match the current requester using facial appearance. Use recorded interactions and visual evidence to distinguish same-category items. Memory may be wrong; check the current scene and correct conflicts. Localize all matching instances separately. Return an empty array if the requested items are absent or unsupported. Follow the query's JSON output format.
\end{vopidRoleCard}

\begin{vopidRoleCard}{User}
Historical person and item registry:\allowbreak{}\par
[\allowbreak{}Stage 1 JSON registry]\allowbreak{}\par
Reference face for video person [\allowbreak{}person ID]\allowbreak{}.\par
[\allowbreak{}registered portrait; repeat for each person]\allowbreak{}\par
Supporting video frame [\allowbreak{}frame ID]\allowbreak{} at [\allowbreak{}timestamp]\allowbreak{} seconds; links:\allowbreak{} [\allowbreak{}item and event references]\allowbreak{}\par
[\allowbreak{}event support frame; repeat for each supplied frame]\allowbreak{}\par
Current localization query. The following two images are Image A and Image B.\par
Image A:\allowbreak{} current requester.\par
[\allowbreak{}requester portrait]\allowbreak{}\par
Image B:\allowbreak{} current scene.\par
[\allowbreak{}numbered test image]\allowbreak{}\par
LOCALIZE. Image A is the current requester's portrait. Image B is the
current scene. Match Image A to video people using faces and the supplied memory;
identify the requested instances through supported person-item interactions.\par
Reference face IDs and video frame IDs are not test-image proposal tags.\par
Use all relevant evidence and preserve the complete original question.\par
Original question:\allowbreak{} [\allowbreak{}full FUN query]\allowbreak{}
Image B has fixed numbered candidate proposals. Select ALL matching
allowed tags,\allowbreak{} each at most once; do not invent tags or coordinates. Output exactly
\{\allowbreak{}"selected\_\allowbreak{}tags":\allowbreak{} [\allowbreak{}1,\allowbreak{} 2]\allowbreak{}\}\allowbreak{} or \{\allowbreak{}"selected\_\allowbreak{}tags":\allowbreak{} [\allowbreak{}]\allowbreak{}\}\allowbreak{}. No additional fields.\par
Allowed tags:\allowbreak{} [\allowbreak{}valid tag IDs]\allowbreak{}
\end{vopidRoleCard}

\vopidFullTitle{ImplicitMem-Box and ImplicitMem-Tag: shared Stage~1}
\begin{vopidRoleCard}{System}
You are a robotic visual assistant. First observe an interaction video and remember people,\allowbreak{} their interactions with distinct physical items,\allowbreak{} item appearances and final locations. In the video-understanding turn output only the requested natural-language memory. In later localization turns use the video and your memory to answer each independent query,\allowbreak{} following that query's JSON output contract.
\end{vopidRoleCard}
\begin{vopidRoleCard}{User}
Named reference portraits are supplied. Associate video people with those names using facial evidence. Preserve uncertain matches as uncertain.\par
Reference portrait [\allowbreak{}registered ID]\allowbreak{}.\par
[\allowbreak{}registered portrait; repeat for each registered user]\allowbreak{}\par
Video [\allowbreak{}frame ID]\allowbreak{} at [\allowbreak{}timestamp]\allowbreak{} seconds.\par
[\allowbreak{}sampled frame; repeat in chronological order]\allowbreak{}\par
Analyze the entire supplied video sequence. Describe every observed person's stable facial features,\allowbreak{} ignoring clothing for identity matching. Record distinct physical items each person visibly wears,\allowbreak{} holds,\allowbreak{} uses,\allowbreak{} touches,\allowbreak{} picks up or places. For each item preserve its category,\allowbreak{} color,\allowbreak{} discriminative appearance,\allowbreak{} interacting person,\allowbreak{} interaction evidence and final location relative to fixed scene landmarks. Keep same-category instances separate and merge repeated observations only when they are clearly the same physical item. Do not infer ownership from proximity. State uncertainty when faces or interactions are unclear. Write a complete natural-language video memory,\allowbreak{} using about 400-800 words if needed. Do not output JSON,\allowbreak{} a registry schema,\allowbreak{} bounding boxes,\allowbreak{} event IDs,\allowbreak{} markdown,\allowbreak{} reasoning or a preamble.
\end{vopidRoleCard}

\par\addvspace{8pt}
\noindent\textbf{ImplicitMem at query time.} Each FUN query starts a separate second turn from the same Stage~1 dialogue. Its context includes the original system instruction, the first user message with the labeled registration profile images and sampled video frames, and the assistant's video description. The two templates below show only the user message appended for the current query. Other queries and their answers are excluded.

\vopidFullTitle{ImplicitMem-Box: Stage~2}
\begin{vopidRoleCard}{User (second turn)}
Locate the physical items requested in the final query. Match the current requester using facial appearance. Use recorded interactions and visual evidence to distinguish same-category items. Memory may be wrong; check the current scene and correct conflicts. Localize all matching instances separately. Return an empty array if the requested items are absent or unsupported. Follow the query's JSON output format.\par
Image A:\allowbreak{} current requester.\par
[\allowbreak{}requester portrait]\allowbreak{}\par
Image B:\allowbreak{} current scene.\par
[\allowbreak{}test image]\allowbreak{}\par
LOCALIZE. Image A is the current requester's portrait. Image B is the current scene. Match Image A to video people using faces and the supplied memory; identify the requested instances through supported person-item interactions.\par
Reference face IDs and video frame IDs are not test-image proposal tags.\par
Use all relevant evidence and preserve the complete original question.\par
Original question:\allowbreak{} [\allowbreak{}full FUN query]\allowbreak{}
Image B has no proposal tags. Return separate tight boxes for ALL requested instances,\allowbreak{} not a union box. Use integer coordinates [\allowbreak{}x1,\allowbreak{}y1,\allowbreak{}x2,\allowbreak{}y2]\allowbreak{} normalized to 0-1000 relative to Image B. Output exactly \{\allowbreak{}"bboxes":\allowbreak{} [\allowbreak{}[\allowbreak{}x1,\allowbreak{}y1,\allowbreak{}x2,\allowbreak{}y2]\allowbreak{},\allowbreak{} ...]\allowbreak{}\}\allowbreak{} or \{\allowbreak{}"bboxes":\allowbreak{} [\allowbreak{}]\allowbreak{}\}\allowbreak{}. No additional fields.
\end{vopidRoleCard}

\vopidFullTitle{ImplicitMem-Tag: Stage~2}
\begin{vopidRoleCard}{User (second turn)}
Image A:\allowbreak{} current requester.\par
[\allowbreak{}requester portrait]\allowbreak{}\par
Image B:\allowbreak{} current scene.\par
[\allowbreak{}numbered test image]\allowbreak{}\par
LOCALIZE. Image A is the current requester's portrait. Image B is the current scene. Match Image A to video people using faces and the supplied memory; identify the requested instances through supported person-item interactions.\par
Reference face IDs and video frame IDs are not test-image proposal tags.\par
Use all relevant evidence and preserve the complete original question.\par
Original question:\allowbreak{} [\allowbreak{}full FUN query]\allowbreak{}
Image B has fixed numbered candidate proposals. Select ALL matching allowed tags,\allowbreak{} each at most once; do not invent tags or coordinates. Output exactly \{\allowbreak{}"selected\_\allowbreak{}tags":\allowbreak{} [\allowbreak{}1,\allowbreak{} 2]\allowbreak{}\}\allowbreak{} or \{\allowbreak{}"selected\_\allowbreak{}tags":\allowbreak{} [\allowbreak{}]\allowbreak{}\}\allowbreak{}. No additional fields.\par
Allowed tags:\allowbreak{} [\allowbreak{}valid tag IDs]\allowbreak{}
\end{vopidRoleCard}

\subsubsection{User Auto-registration (UAR)}
\label{supplsec:prompt-uar}
Stage~1 receives the reference video without registered profile images or a list of identities. Stage~2 receives the requester's profile image with each FUN query.

\vopidFullTitle{ExplicitMem-Box and ExplicitMem-Tag: shared Stage~1}
\begin{vopidRoleCard}{System}
You are a visual assistant for personal-item localization.\par
Identify people using stable facial evidence,\allowbreak{} not clothing. Track distinct physical
items through visible holding,\allowbreak{} wearing,\allowbreak{} use,\allowbreak{} touch,\allowbreak{} pickup and placement. Proximity
alone does not establish a person-item association. Keep same-category instances
separate. Preserve explicit attributes in the original question. Do not invent
identity,\allowbreak{} interaction or ownership evidence. Inspect the current scene rather than
assuming an item's last video location is unchanged. Return every requested visible
physical instance separately; do not include another person's items,\allowbreak{} duplicate
instances or unsupported extras. Return an explicit empty set when unsupported.\par
No thinking,\allowbreak{} markdown or text outside the requested JSON object.\par
There are two tasks:\allowbreak{} BUILD\_\allowbreak{}MEMORY before any localization question,\allowbreak{} and LOCALIZE.\par
For BUILD\_\allowbreak{}MEMORY output the requested registry schema; for LOCALIZE output only
the requested localization schema. Historical memory is predicted evidence,\allowbreak{} not GT.\par
IDENTITY CONDITION:\allowbreak{} During BUILD\_\allowbreak{}MEMORY,\allowbreak{} no reference portraits or list of video people is supplied.\par
Discover people only from the video frames. Assign local IDs P001,\allowbreak{} P002,\allowbreak{} ...;
do not infer names. For each registered person choose one supplied frame where
their face is clearly visible and return face\_\allowbreak{}reference with that exact frame\_\allowbreak{}id
and a tight face bbox [\allowbreak{}x1,\allowbreak{}y1,\allowbreak{}x2,\allowbreak{}y2]\allowbreak{} on a 0-1000 grid relative to the whole frame.\par
Use stable faces to maintain IDs; never use query portraits or queries in this task.\par
Register only people with a supported face reference; uncertain item links use an
empty person ID/list. Do not invent a face or a frame.
\end{vopidRoleCard}
\begin{vopidRoleCard}{User}
Video [\allowbreak{}frame ID]\allowbreak{} at [\allowbreak{}timestamp]\allowbreak{} seconds.\par
[\allowbreak{}sampled frame; repeat in chronological order]\allowbreak{}\par
BUILD\_\allowbreak{}MEMORY. Observe the entire supplied timestamped video sequence.\par
Output a people registry and an items registry. Describe each person's stable face.\par
Register at most 12 visibly interacted item instances,\allowbreak{} merge repeated observations,\allowbreak{}
and preserve their discriminative appearance and final location relative to fixed
scene landmarks. Each item has one or two evidence events:\allowbreak{} an informative
interaction and,\allowbreak{} when observed,\allowbreak{} final placement. Use actual printed seconds from
the video start,\allowbreak{} unique event IDs,\allowbreak{} and only registered person IDs or an empty ID.\par
Keep string fields concise. Output exactly the supplied JSON schema.\par
Output JSON schema:\allowbreak{}\par
[\allowbreak{}registry schema for this setting and video]\allowbreak{}
\end{vopidRoleCard}

\vopidFullTitle{ExplicitMem-Box: Stage~2}
\begin{vopidRoleCard}{System}
Locate the physical items requested in the final query. Match the current requester using facial appearance. Use recorded interactions and visual evidence to distinguish same-category items. Memory may be wrong; check the current scene and correct conflicts. Localize all matching instances separately. Return an empty array if the requested items are absent or unsupported. Follow the query's JSON output format.
\end{vopidRoleCard}
\begin{vopidRoleCard}{User}
Historical person and item registry:\allowbreak{}\par
[\allowbreak{}Stage 1 JSON registry]\allowbreak{}\par
Reference face for video person [\allowbreak{}person ID]\allowbreak{}.\par
[\allowbreak{}video face crop; repeat for each person]\allowbreak{}\par
Supporting video frame [\allowbreak{}frame ID]\allowbreak{} at [\allowbreak{}timestamp]\allowbreak{} seconds; links:\allowbreak{} [\allowbreak{}item and event references]\allowbreak{}\par
[\allowbreak{}event support frame; repeat for each supplied frame]\allowbreak{}\par
Current localization query. The following two images are Image A and Image B.\par
Image A:\allowbreak{} current requester.\par
[\allowbreak{}requester portrait]\allowbreak{}\par
Image B:\allowbreak{} current scene.\par
[\allowbreak{}test image]\allowbreak{}\par
LOCALIZE. Image A is the current requester's portrait. Image B is the
current scene. Match Image A to video people using faces and the supplied memory;
identify the requested instances through supported person-item interactions.\par
Reference face IDs and video frame IDs are not test-image proposal tags.\par
Use all relevant evidence and preserve the complete original question.\par
Original question:\allowbreak{} [\allowbreak{}full FUN query]\allowbreak{}
Image B has no proposal tags. Return separate tight boxes for ALL requested
instances,\allowbreak{} not a union box. Use integer coordinates [\allowbreak{}x1,\allowbreak{}y1,\allowbreak{}x2,\allowbreak{}y2]\allowbreak{} normalized to
0-1000 relative to Image B. Output exactly \{\allowbreak{}"bboxes":\allowbreak{} [\allowbreak{}[\allowbreak{}x1,\allowbreak{}y1,\allowbreak{}x2,\allowbreak{}y2]\allowbreak{},\allowbreak{} ...]\allowbreak{}\}\allowbreak{} or
\{\allowbreak{}"bboxes":\allowbreak{} [\allowbreak{}]\allowbreak{}\}\allowbreak{}. No additional fields.
\end{vopidRoleCard}

\vopidFullTitle{ExplicitMem-Tag: Stage~2}
\begin{vopidRoleCard}{System}
Locate the physical items requested in the final query. Match the current requester using facial appearance. Use recorded interactions and visual evidence to distinguish same-category items. Memory may be wrong; check the current scene and correct conflicts. Localize all matching instances separately. Return an empty array if the requested items are absent or unsupported. Follow the query's JSON output format.
\end{vopidRoleCard}
\begin{vopidRoleCard}{User}
Historical person and item registry:\allowbreak{}\par
[\allowbreak{}Stage 1 JSON registry]\allowbreak{}\par
Reference face for video person [\allowbreak{}person ID]\allowbreak{}.\par
[\allowbreak{}video face crop; repeat for each person]\allowbreak{}\par
Supporting video frame [\allowbreak{}frame ID]\allowbreak{} at [\allowbreak{}timestamp]\allowbreak{} seconds; links:\allowbreak{} [\allowbreak{}item and event references]\allowbreak{}\par
[\allowbreak{}event support frame; repeat for each supplied frame]\allowbreak{}\par
Current localization query. The following two images are Image A and Image B.\par
Image A:\allowbreak{} current requester.\par
[\allowbreak{}requester portrait]\allowbreak{}\par
Image B:\allowbreak{} current scene.\par
[\allowbreak{}numbered test image]\allowbreak{}\par
LOCALIZE. Image A is the current requester's portrait. Image B is the
current scene. Match Image A to video people using faces and the supplied memory;
identify the requested instances through supported person-item interactions.\par
Reference face IDs and video frame IDs are not test-image proposal tags.\par
Use all relevant evidence and preserve the complete original question.\par
Original question:\allowbreak{} [\allowbreak{}full FUN query]\allowbreak{}
Image B has fixed numbered candidate proposals. Select ALL matching
allowed tags,\allowbreak{} each at most once; do not invent tags or coordinates. Output exactly
\{\allowbreak{}"selected\_\allowbreak{}tags":\allowbreak{} [\allowbreak{}1,\allowbreak{} 2]\allowbreak{}\}\allowbreak{} or \{\allowbreak{}"selected\_\allowbreak{}tags":\allowbreak{} [\allowbreak{}]\allowbreak{}\}\allowbreak{}. No additional fields.\par
Allowed tags:\allowbreak{} [\allowbreak{}valid tag IDs]\allowbreak{}
\end{vopidRoleCard}

\vopidFullTitle{ImplicitMem-Box and ImplicitMem-Tag: shared Stage~1}
\begin{vopidRoleCard}{System}
You are a robotic visual assistant. First observe an interaction video and remember people,\allowbreak{} their interactions with distinct physical items,\allowbreak{} item appearances and final locations. In the video-understanding turn output only the requested natural-language memory. In later localization turns use the video and your memory to answer each independent query,\allowbreak{} following that query's JSON output contract.
\end{vopidRoleCard}
\begin{vopidRoleCard}{User}
No reference portraits are supplied. Describe and distinguish people using the video alone. Do not invent real names.\par
Video [\allowbreak{}frame ID]\allowbreak{} at [\allowbreak{}timestamp]\allowbreak{} seconds.\par
[\allowbreak{}sampled frame; repeat in chronological order]\allowbreak{}\par
Analyze the entire supplied video sequence. Describe every observed person's stable facial features,\allowbreak{} ignoring clothing for identity matching. Record distinct physical items each person visibly wears,\allowbreak{} holds,\allowbreak{} uses,\allowbreak{} touches,\allowbreak{} picks up or places. For each item preserve its category,\allowbreak{} color,\allowbreak{} discriminative appearance,\allowbreak{} interacting person,\allowbreak{} interaction evidence and final location relative to fixed scene landmarks. Keep same-category instances separate and merge repeated observations only when they are clearly the same physical item. Do not infer ownership from proximity. State uncertainty when faces or interactions are unclear. Write a complete natural-language video memory,\allowbreak{} using about 400-800 words if needed. Do not output JSON,\allowbreak{} a registry schema,\allowbreak{} bounding boxes,\allowbreak{} event IDs,\allowbreak{} markdown,\allowbreak{} reasoning or a preamble.
\end{vopidRoleCard}

\par\addvspace{8pt}
\noindent\textbf{ImplicitMem at query time.} Each FUN query starts a separate second turn from the same Stage~1 dialogue. Its context includes the original system instruction, the first user message with the sampled video frames, and the assistant's video description. The two templates below show only the user message appended for the current query. Other queries and their answers are excluded.

\vopidFullTitle{ImplicitMem-Box: Stage~2}
\begin{vopidRoleCard}{User (second turn)}
Locate the physical items requested in the final query. Match the current requester using facial appearance. Use recorded interactions and visual evidence to distinguish same-category items. Memory may be wrong; check the current scene and correct conflicts. Localize all matching instances separately. Return an empty array if the requested items are absent or unsupported. Follow the query's JSON output format.\par
Image A:\allowbreak{} current requester.\par
[\allowbreak{}requester portrait]\allowbreak{}\par
Image B:\allowbreak{} current scene.\par
[\allowbreak{}test image]\allowbreak{}\par
LOCALIZE. Image A is the current requester's portrait. Image B is the current scene. Match Image A to video people using faces and the supplied memory; identify the requested instances through supported person-item interactions.\par
Reference face IDs and video frame IDs are not test-image proposal tags.\par
Use all relevant evidence and preserve the complete original question.\par
Original question:\allowbreak{} [\allowbreak{}full FUN query]\allowbreak{}
Image B has no proposal tags. Return separate tight boxes for ALL requested instances,\allowbreak{} not a union box. Use integer coordinates [\allowbreak{}x1,\allowbreak{}y1,\allowbreak{}x2,\allowbreak{}y2]\allowbreak{} normalized to 0-1000 relative to Image B. Output exactly \{\allowbreak{}"bboxes":\allowbreak{} [\allowbreak{}[\allowbreak{}x1,\allowbreak{}y1,\allowbreak{}x2,\allowbreak{}y2]\allowbreak{},\allowbreak{} ...]\allowbreak{}\}\allowbreak{} or \{\allowbreak{}"bboxes":\allowbreak{} [\allowbreak{}]\allowbreak{}\}\allowbreak{}. No additional fields.
\end{vopidRoleCard}

\vopidFullTitle{ImplicitMem-Tag: Stage~2}
\begin{vopidRoleCard}{User (second turn)}
Image A:\allowbreak{} current requester.\par
[\allowbreak{}requester portrait]\allowbreak{}\par
Image B:\allowbreak{} current scene.\par
[\allowbreak{}numbered test image]\allowbreak{}\par
LOCALIZE. Image A is the current requester's portrait. Image B is the current scene. Match Image A to video people using faces and the supplied memory; identify the requested instances through supported person-item interactions.\par
Reference face IDs and video frame IDs are not test-image proposal tags.\par
Use all relevant evidence and preserve the complete original question.\par
Original question:\allowbreak{} [\allowbreak{}full FUN query]\allowbreak{}
Image B has fixed numbered candidate proposals. Select ALL matching allowed tags,\allowbreak{} each at most once; do not invent tags or coordinates. Output exactly \{\allowbreak{}"selected\_\allowbreak{}tags":\allowbreak{} [\allowbreak{}1,\allowbreak{} 2]\allowbreak{}\}\allowbreak{} or \{\allowbreak{}"selected\_\allowbreak{}tags":\allowbreak{} [\allowbreak{}]\allowbreak{}\}\allowbreak{}. No additional fields.\par
Allowed tags:\allowbreak{} [\allowbreak{}valid tag IDs]\allowbreak{}
\end{vopidRoleCard}

\subsubsection{Modern Baseline}
\label{supplsec:prompt-modern}
The modern baseline uses the same single request under UPR and UAR. It receives the sampled reference frames and the current query together and predicts boxes directly.

\vopidFullTitle{Modern baseline: single request}
\begin{vopidRoleCard}{System}
Locate the physical items requested in the final query. Match the current requester using facial appearance. Use recorded interactions and visual evidence to distinguish same-category items. Memory may be wrong; check the current scene and correct conflicts. Localize all matching instances separately. Return an empty array if the requested items are absent or unsupported. Follow the query's JSON output format.
\end{vopidRoleCard}
\begin{vopidRoleCard}{User}
Historical video frames,\allowbreak{} in chronological order.\par
Video [\allowbreak{}frame ID]\allowbreak{} at [\allowbreak{}timestamp]\allowbreak{} seconds.\par
[\allowbreak{}sampled frame; repeat in chronological order]\allowbreak{}\par
Current localization query. The following two images are Image A and Image B.\par
Image A:\allowbreak{} current requester.\par
[\allowbreak{}requester portrait]\allowbreak{}\par
Image B:\allowbreak{} current scene.\par
[\allowbreak{}test image]\allowbreak{}\par
LOCALIZE. Image A is the current requester's portrait. Image B is the\par
current scene. Match Image A to video people using faces and the supplied memory;\par
identify the requested instances through supported person-item interactions.\par
Reference face IDs and video frame IDs are not test-image proposal tags.\par
Use all relevant evidence and preserve the complete original question.\par
Original question:\allowbreak{} [\allowbreak{}full FUN query]\allowbreak{}
Image B has no proposal tags. Return separate tight boxes for ALL requested
instances,\allowbreak{} not a union box. Use integer coordinates [\allowbreak{}x1,\allowbreak{}y1,\allowbreak{}x2,\allowbreak{}y2]\allowbreak{} normalized to
0-1000 relative to Image B. Output exactly \{\allowbreak{}"bboxes":\allowbreak{} [\allowbreak{}[\allowbreak{}x1,\allowbreak{}y1,\allowbreak{}x2,\allowbreak{}y2]\allowbreak{},\allowbreak{} ...]\allowbreak{}\}\allowbreak{} or
\{\allowbreak{}"bboxes":\allowbreak{} [\allowbreak{}]\allowbreak{}\}\allowbreak{}. No additional fields.
\end{vopidRoleCard}

\subsubsection{Target-Phrase Extraction for Tag Methods}
\label{supplsec:prompt-phrase-blocks}
During Stage~2 of both Tag methods, Qwen3.5-0.8B extracts the phrase used to prompt LocateAnything. The parser receives the question text alone. Its system message includes the examples below.

\vopidFullTitle{Target-phrase extraction: initial request}
\begin{vopidRoleCard}{System}
Extract the physical target object from a robot item-finding command. Return JSON only with keys category,\allowbreak{} attributes,\allowbreak{} prompt. category must be the concrete object noun,\allowbreak{} not a broad class. attributes must be a JSON array of short visual attributes useful for detection. Normalize ordinary object synonyms when helpful,\allowbreak{} for example earphone to headphones,\allowbreak{} mug to cup,\allowbreak{} cellphone to phone,\allowbreak{} or sneakers to shoes. Do not add unsupported colors,\allowbreak{} brands,\allowbreak{} or materials. prompt is passed directly to LocateAnything,\allowbreak{} so it must be a short English noun phrase,\allowbreak{} not a sentence. Intelligently retain an explicitly written visual modifier when it helps distinguish the target,\allowbreak{} but omit modifiers that are unlikely to help proposal recall. Do not include person identity,\allowbreak{} ownership,\allowbreak{} location,\allowbreak{} instructions,\allowbreak{} or a complete sentence in prompt. Examples:\allowbreak{} \{\allowbreak{}"category":\allowbreak{}"headphones",\allowbreak{}"attributes":\allowbreak{}[\allowbreak{}"white"]\allowbreak{},\allowbreak{}"prompt":\allowbreak{}"white headphones"\}\allowbreak{}; \{\allowbreak{}"category":\allowbreak{}"shoes",\allowbreak{}"attributes":\allowbreak{}[\allowbreak{}"NIKE"]\allowbreak{},\allowbreak{}"prompt":\allowbreak{}"shoes"\}\allowbreak{}; \{\allowbreak{}"category":\allowbreak{}"boots",\allowbreak{}"attributes":\allowbreak{}[\allowbreak{}"black"]\allowbreak{},\allowbreak{}"prompt":\allowbreak{}"boots"\}\allowbreak{}; \{\allowbreak{}"category":\allowbreak{}"cup",\allowbreak{}"attributes":\allowbreak{}[\allowbreak{}]\allowbreak{},\allowbreak{}"prompt":\allowbreak{}"cup"\}\allowbreak{}; \{\allowbreak{}"category":\allowbreak{}"mug",\allowbreak{}"attributes":\allowbreak{}[\allowbreak{}"white"]\allowbreak{},\allowbreak{}"prompt":\allowbreak{}"white mug"\}\allowbreak{}; \{\allowbreak{}"category":\allowbreak{}"mug",\allowbreak{}"attributes":\allowbreak{}[\allowbreak{}"patterned"]\allowbreak{},\allowbreak{}"prompt":\allowbreak{}"patterned mug"\}\allowbreak{}; \{\allowbreak{}"category":\allowbreak{}"laptop",\allowbreak{}"attributes":\allowbreak{}[\allowbreak{}"gaming"]\allowbreak{},\allowbreak{}"prompt":\allowbreak{}"gaming laptop"\}\allowbreak{}.
\end{vopidRoleCard}
\begin{vopidRoleCard}{User}
Command:\allowbreak{} [\allowbreak{}full FUN query]\allowbreak{}
Return JSON only,\allowbreak{} e.g. \{\allowbreak{}"category":\allowbreak{}"headphones",\allowbreak{}"attributes":\allowbreak{}[\allowbreak{}"white"]\allowbreak{},\allowbreak{}"prompt":\allowbreak{}"white headphones"\}\allowbreak{}
\end{vopidRoleCard}

\endgroup

\section{Additional Experiments and Visualizations} 
\label{supplsec:more_results}
We examine how effectively ImplicitMem-Tag distinguishes same-category items across different users and how video sampling density affects localization on longer videos. We also present qualitative examples that illustrate the distinction between recognizing an object category and identifying the specific instance intended by the user.

\subsection{Diagnostic Analysis and Sampling Sensitivity}
\label{supplsec:pending-ablations}
We conduct diagnostic experiments using Qwen3.8-27B ImplicitMem-Tag under the UPR setting on the VOPID test set, keeping the original FUN queries and precomputed LocateAnything proposals unchanged.
All metrics follow the evaluation protocol in \S\ref{supplsec:evaluation}.

\begin{wraptable}{r}{0.48\linewidth}
\centering\small
\setlength{\tabcolsep}{3pt}
\vspace{-17mm}
\caption{\small \textbf{Same-category requests from different users.} We report query counts and mAP (\%) for ImplicitMem-Tag under UPR. Only target-present queries are included.}
\label{tab:supp-multi-requester}
\begin{tabular*}{\linewidth}{@{\extracolsep{\fill}}lrr@{}}
\toprule
Test subset & count & mAP (\%) \\
\midrule
same-category queries & 216 & 23.80 \\
remaining queries & 139 & \textbf{42.16} \\
\midrule
all target-present queries & 355 & 30.99 \\
\bottomrule
\end{tabular*}
\vspace{-3mm}
\end{wraptable}

\subsubsection{Same-Category Items Across Users}
\label{supplsec:association-controls}
To assess instance-level ownership disambiguation beyond category recognition, we identify target-present queries for which at least one other user associated with the same reference video asks about an item of the same category. Of the 355 target-present test queries, 216 meet this criterion across 23 reference videos (Table~\ref{tab:supp-dataset-statistics}). The remaining 139 serve as the comparison group.

As shown in Table~\ref{tab:supp-multi-requester}, ImplicitMem-Tag achieves 23.80 $\mathrm{mAP}$ on the same-category subset under UPR, compared with 42.16 on the remaining queries and 30.99 across all 355 target-present queries. The 18.36-point gap suggests that distinguishing belongings of the same category across users remains challenging.

\begin{wraptable}{r}{0.48\linewidth}
\centering
\small
\setlength{\tabcolsep}{6pt}
\renewcommand{\arraystretch}{0.92}
\vspace{-9mm}
\caption{\small \textbf{Effect of the video frame cap under UPR.} \textit{All} and \textit{long} report mAP (\%) for the full test set (389 queries) and for videos longer than two minutes (237 queries), respectively. \textit{KMV} gives the mean time to build memory for one video. \textit{Inference} gives the mean time per query from phrase extraction through proposal generation, tag rendering, and MLLM selection. Both times are in seconds.}
\label{tab:supp-pending-context}
\begin{tabular}{@{}lrrrr@{}}
\toprule
\multirow{2}{*}{Frame cap} & \multirow{2}{*}{KMV $\downarrow$} & \multirow{2}{*}{inference $\downarrow$} & \multicolumn{2}{c}{mAP $\uparrow$} \\
\cmidrule(l){4-5}
& & & all & long \\
\midrule
24 & \textbf{61.93} & \textbf{6.02} & 30.02 & 25.86 \\
48 (default) & 66.55 & 6.14 & 30.33 & 26.62 \\
72 & 71.17 & 6.42 & 30.98 & 26.54 \\
96 & 71.73 & 6.56 & \textbf{31.36} & \textbf{27.93} \\
\bottomrule
\end{tabular}
\vspace{-1mm}
\end{wraptable}

\subsubsection{Video Sampling Budget}
\label{supplsec:context-controls}
To examine how temporal sampling density affects localization, we evaluate Qwen3.8-27B ImplicitMem-Tag under UPR with maximum frame caps ranging from 24 to 96.
Following our uniform sampling scheme at a target rate of 1\,fps, raising the frame cap yields denser temporal coverage on longer videos, whereas shorter videos retain the same sampled frames across budgets.
All runs use identical frame resolution, prompts, candidate proposals, and token budgets for generation.

As reported in Table~\ref{tab:supp-pending-context}, localization accuracy improves steadily with larger frame budgets, peaking at 96 frames (31.36 $\mathrm{mAP}$ overall and 27.93 $\mathrm{mAP}$ on the 237 queries from videos longer than two minutes).
Doubling the default cap from 48 to 96 frames improves overall $\mathrm{mAP}$ by $1.03$ points (30.33 $\rightarrow$ 31.36), with a slightly larger gain of 1.31 points (26.62 $\rightarrow$ 27.93) on long videos.
In terms of runtime, memory construction (\texttt{KMV}) increases by 5.18\,s per video (66.55 $\rightarrow$ 71.73\,s), while mean per-query inference time increases by only 0.42\,s (6.14 $\rightarrow$ 6.56\,s).
These results indicate that the default 48-frame budget already captures most ownership-relevant interactions while offering a favorable balance between accuracy and memory-construction overhead.

\subsection{More Visualizations}
\label{supplsec:qualitative}
\begin{figure}[t]
\includegraphics[width=\textwidth,height=.78\textheight,keepaspectratio]{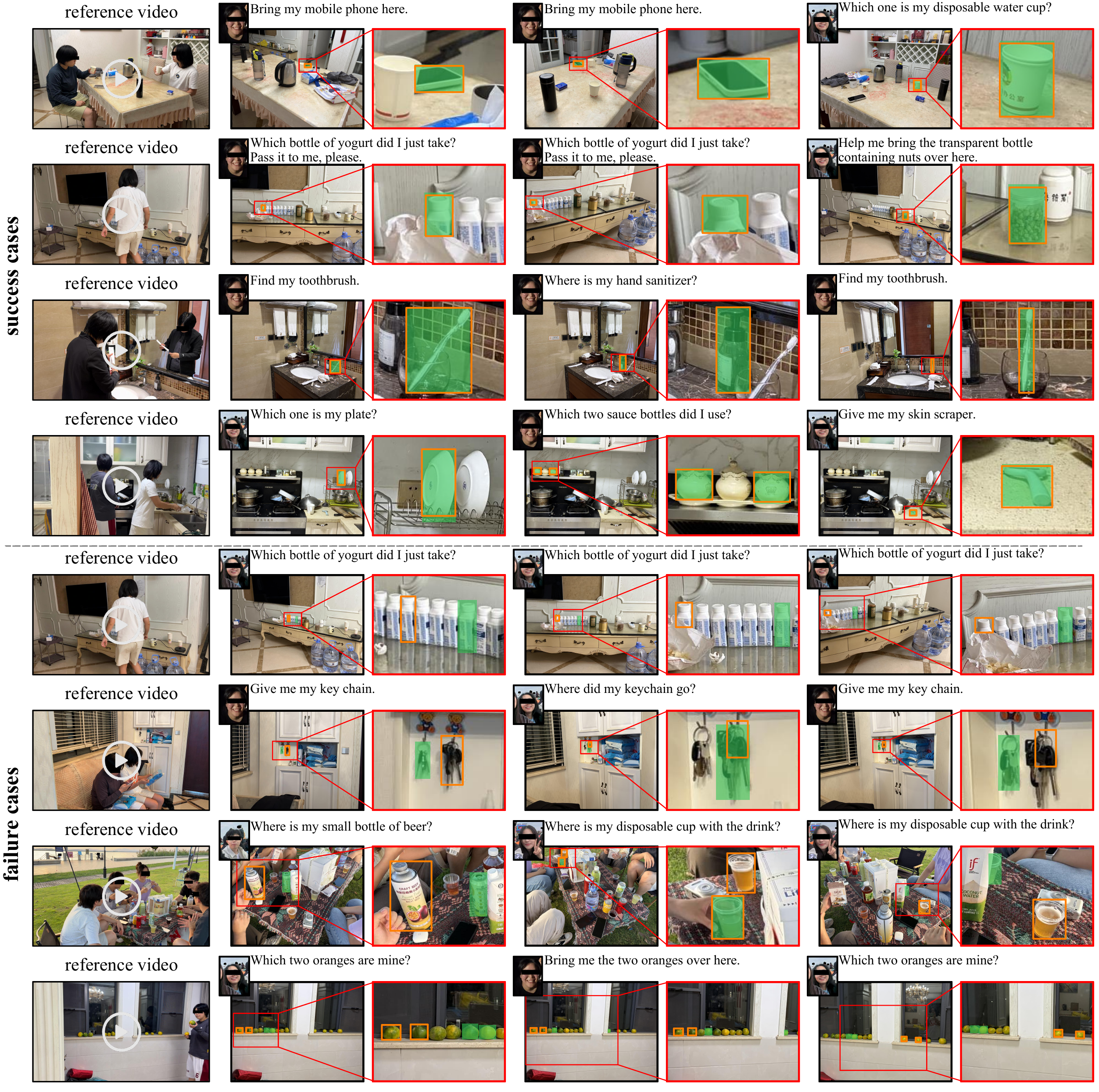}
\vspace{-9mm}
\caption{\small \textbf{Qualitative results of ImplicitMem-Tag under UPR.} Successful examples appear above the dashed line and failures below. Each row shows a reference video frame and three FUN queries, each accompanied by the user's profile image and the test image. Red rectangles indicate the regions shown in the enlarged views. \textcolor{green!70!black}{Green shading} marks the ground-truth boxes, and \textcolor{orange}{orange outlines} mark the model's predictions. Eyes are masked for privacy.}
\label{fig:supp-qualitative-results}
\vspace{-3mm}
\end{figure}

Fig.~\ref{fig:supp-qualitative-results} presents qualitative predictions from Qwen3.8-27B ImplicitMem-Tag under UPR across 24 target-present test queries. On the positive side, the model accurately localizes small personal items such as toothbrushes and successfully retrieves multiple items belonging to the same user, such as two sauce bottles. However, visually similar distractors of the same category can still cause instance-selection errors: in one test image, the model selects an adjacent yogurt bottle instead of the target one, while in another, it erroneously selects an additional cup alongside the user's own cup.

\section{Human Evaluation}
\label{supplsec:human}
\subsection{Access to the Reference Video}
\label{supplsec:human-protocol}
Both human evaluations follow the user auto-registration (UAR) setting. Annotators watch the reference video without pre-registered profile images and receive the user's profile image alongside each FUN query. In the \textit{closed-book} condition, annotators watch each video once and rely solely on memory when localizing targets in the test images. In the \textit{open-book} condition, they are allowed to revisit the video during annotation.

To assess human performance, multiple annotators were independently recruited to label a random set of VOPID videos.
On the VOPID test set, annotators achieve 82.91 $\mathrm{mAP}$ after a single viewing and 91.76 $\mathrm{mAP}$ when allowed to revisit the video.
Both scores substantially outperform Qwen3.8-27B ImplicitMem-Tag (31.23 $\mathrm{mAP}$), the strongest automated method under UAR; even under the closed-book condition, human performance exceeds that of ImplicitMem-Tag by 51.68 points.

\subsection{Annotation and Scoring}
For each FUN query, annotators use the provided profile image to identify the user and rely on interactions observed in the reference video to infer which items belong to that person. They annotate each visible target instance in the test image with a separate bounding box. When multiple objects match the query description, ownership determines which instances to annotate. If none of the requested items is visible, annotators return an empty set.

Human annotations follow the visibility and bounding-box conventions in \S\ref{supplsec:annotation}, and are scored using the same query-level matching protocol as the automated methods detailed in \S\ref{supplsec:evaluation}.

\section{Code Availability and Experimental Environment}
\label{supplsec:code}
\textbf{Code and data availability.} The supplementary materials include an anonymized code package for data preprocessing, memory construction, query inference, and evaluation. The package also provides setup instructions, prompt templates, and the configurations used for the main PID experiments. We omit repository links during anonymous review. The submission package excludes the VOPID dataset because of its size and the need for controlled access to images of identifiable participants.

\textbf{Experimental environment.} We run MLLM inference in BF16 with two-way tensor parallelism across two NVIDIA RTX A6000 GPUs, each with 48\,GB of memory. The software environment uses PyTorch~2.10.0, vLLM~0.19.1, and Transformers~5.5.4.
\S\ref{supplsec:sampling} details the model configurations, decoding parameters, and video preprocessing procedure.




\end{document}